\documentclass[11pt]{article}
\usepackage[a4paper,margin=1in]{geometry}
\usepackage{graphicx}
\usepackage{subcaption}
\usepackage{booktabs}
\usepackage{amssymb}
\usepackage{amsmath}
\usepackage{pifont}
\newcommand{\xmark}{\ding{55}}
\usepackage{tabularx}
\usepackage{multirow}
\usepackage{hyperref}
\hypersetup{hidelinks}
\usepackage[super,sort&compress]{natbib}
\usepackage{placeins}
\usepackage{authblk}
\title{RaspAssistant as an Agentic Framework for Automating Adsorption Simulation Research in Porous Materials}
\author[1,3]{Marko Petkovi\'c}
\author[1]{Susanne Gooijer}
\author[1]{Jos\'e Manuel Vicent-Luna}
\author[2,3]{Vlado Menkovski}
\author[1,3]{Sof\'ia Calero\thanks{Corresponding author: \href{mailto:s.calero@tue.nl}{s.calero@tue.nl}}}
\affil[1]{Department of Applied Physics and Science Education, Eindhoven University of Technology, Eindhoven, The Netherlands}
\affil[2]{Department of Mathematics and Computer Science, Eindhoven University of Technology, Eindhoven, The Netherlands}
\affil[3]{Eindhoven Artificial Intelligence Systems Institute (EAISI), Eindhoven University of Technology, Eindhoven, The Netherlands}
\date{}

\usepackage{fvextra}
\DefineVerbatimEnvironment{BenchmarkPrompt}{Verbatim}{breaklines=true,breakanywhere=true,breaksymbolleft={},fontsize=\small}
\makeatletter
\let\PreprintMakeTitle\maketitle
\let\PreprintInnerTitle\@maketitle
\let\PreprintThanks\thanks
\let\PreprintAuthor\@author
\makeatother
\begin{document}

\maketitle

\begin{abstract}
Porous-material adsorption studies require coordinated structure preparation, force-field selection, simulation, validation, and interpretation, making end-to-end automation difficult. We test whether RaspAssistant, a hierarchical multi-agent framework, can carry an adsorption study from force-field preparation through validation and downstream separation analysis. Across twelve multi-agent workflow-reproduction runs spanning four adsorption and separation benchmarks, overall scores ranged from 8.50 to 10.00 on a rubric evaluating workflow correctness, simulation validity, and analysis consistency. In a separate eight-case force-field extraction benchmark, the mean weighted score was 0.967, combining parameter agreement and RASPA readiness. In open-ended studies, the system diagnosed and refined an acetylene force field and, after validating a CO$_2$ force field against MFI experiments, screened 197 pure-silica zeolites that passed automated input checks. These results demonstrate autonomous workflow execution after task specification and input provision: RaspAssistant coordinates, validates, adapts, and documents extended adsorption-simulation workflows while exposing model sensitivity and methodological limitations for expert review.
\end{abstract}

\section{Introduction}
Molecular simulations connect atomistic structure to macroscopic performance, reveal mechanisms often difficult to probe experimentally, and evaluate candidate materials before synthesis and characterization. For porous materials, adsorption and separation depend sensitively on framework topology, pore chemistry, and host--guest interactions. Packages such as RASPA \cite{dubbeldam2016raspa,ran2024raspa3} evaluate adsorption, diffusion, and mixture behaviour in nanoporous materials, while computation-ready databases such as CoRE MOF \cite{chung2019coremof,zhao2025coremofdb} support systematic screening \cite{ongari2020automation}.

Despite mature simulation tools, reliable porous-materials studies remain complex and expertise-intensive. Researchers must prepare computation-ready structures, choose supercells and simulation conditions, assign or adapt framework and adsorbate force fields, and decide whether transferable models such as UFF4MOF \cite{addicoat2014uff4mof,coupry2016uff4mofext}, DREIDING \cite{mayo1990dreiding}, or TraPPE \cite{martin1998trappe} need system-specific refinement. MOF adsorption predictions can depend strongly on force-field selection \cite{mccready2024ffuncertainty}, while experimental variability and occasional outliers complicate validation \cite{park2017reproducibility,bingel2022alkane}.

An adsorption and separation study extends beyond a single simulation, coupling pure-component isotherms, experimental validation, mixture predictions using Ideal Adsorbed Solution Theory (IAST) or direct simulations, and breakthrough analysis. Tools support isotherm fitting, mixture prediction, and process simulation \cite{simon2016pyiast,sharma2023ruptura,hassan2026aim}. Complete automation also requires coordinating these stages and revising models or protocols when validation exposes limitations.

These coordination needs motivate agentic scientific systems, which combine large language models (LLMs) with planning, memory, tool use, and iterative reasoning \cite{dral2026machine}. Such systems can translate research objectives into tool-based actions and use the results to guide subsequent steps. Applications span experimental chemistry \cite{boiko2023coscientist,bran2024chemcrow}, quantum chemistry \cite{zou2025agente,pham2026chemgraph}, biomolecular simulation \cite{campbell2025mdcrow}, atomistic materials modeling \cite{wang2025dreams,liu2025vaspilot,chaudhari2026matsciagent,soleymanibrojeni2026genius,vriza2026atomistic,rothfarb2026master,liu2026masgent,somasundaram2026ursa,huang2026alkemie,yang2026quasar,ghafarollahi2025atomagents,shi2025mdagent}, and porous-material property prediction and design \cite{kang2024chatmof,lee2026simmof}.

\begin{figure}[!htbp]
\centering
\includegraphics[width=0.95\linewidth]{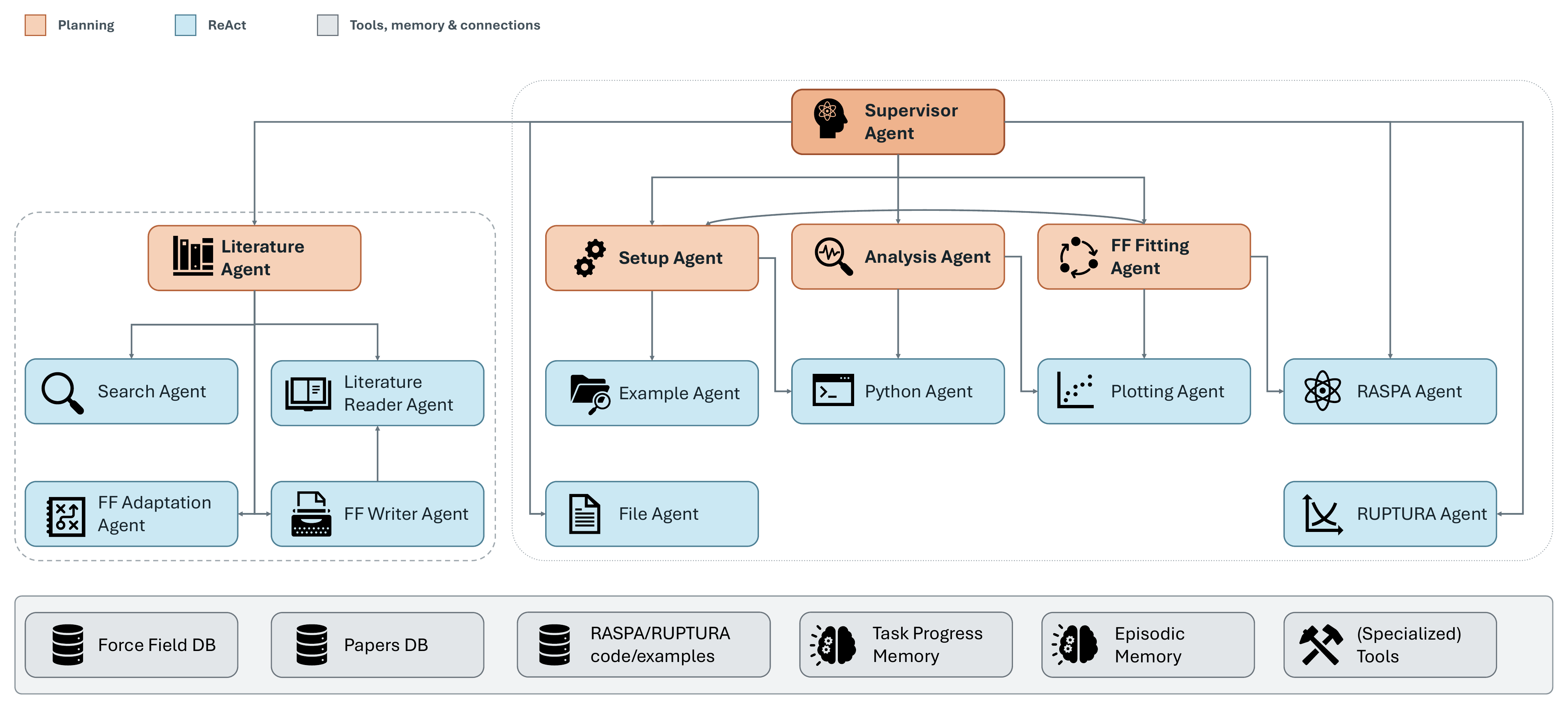}
\caption{Agent architecture of RaspAssistant. Orange nodes denote higher-level agents responsible for planning, coordination, and verification, while blue nodes denote lower-level ReAct agents that alternate reasoning with tool actions. The red dashed subsystem supports literature search, force field extraction, adaptation, and parameter file generation. The blue dashed subsystem supports simulation setup, execution, analysis, plotting, and force field fitting for workflow reproduction. Shared databases, examples, memories, and tools occupy the bottom layer. Agent roles are listed in Table~\ref{SI-tab:agents}.}
\label{fig:raspassist}
\end{figure}

We introduce RaspAssistant, a multi-agent framework (Figure~\ref{fig:raspassist}) for literature-informed force-field construction, input preparation, simulation, post-processing, and analysis across adsorption and separation workflows. We test whether it can reproduce a complete workflow without human scientific intervention after task specification and input preparation. Building on our setup-focused framework \cite{petkovic2025fullyautomated}, agents use complete simulation and force-field files, general file-manipulation tools, and execution tools to revise setup and task sequences as evidence accumulates. We evaluate four published workflows, literature-based force-field extraction, and two open-ended studies of model refinement and validation-based zeolite screening. The contribution is continuity across force-field selection or adaptation, experimental validation, pure-component and mixture simulation, and process-level analysis. Further context is in the Supporting Information (SI).

\FloatBarrier

\section{Results}
\subsection{Workflow Reproduction}
The benchmark comprises four published studies: (1) CO$_2$/N$_2$ separation in TAMOF-1 \cite{gooijer2025tamof}, (2) C$_2$H$_2$/CO$_2$ and C$_2$H$_2$/CH$_4$ separation in open-metal-site MOFs \cite{luna2019acetylene}, (3) metal-center effects in M-MOF-74 \cite{wasik2024impact}, and (4) C$_3$--C$_6$ olefin/paraffin separation in cation-exchanged zeolites \cite{luna2020pi}. Literature-derived force fields were supplied for these runs. Each study had three multi-agent runs scored for workflow correctness, simulation validity, and analysis consistency (methods in the SI), plus one single-agent baseline reported separately and excluded from multi-agent averages. Table~\ref{tab:workflow_resources} lists run-level scores and approximate resource use.

\begin{table}[htbp]
\centering
\caption{Scores and approximate resource use for the twelve multi-agent workflow-reproduction runs.}
\label{tab:workflow_resources}
\small
\begin{tabular}{llrrrr}
\hline
\textbf{Benchmark} & \textbf{Run} & \textbf{Overall (/10)} & \textbf{Elapsed (d)} & \textbf{Tokens (M)} & \textbf{Cost (US\$)} \\
\hline
TAMOF-1 & 1 & 10.00 & 8.8 & 12 & 15 \\
 & 2 & 9.60 & 2.4 & 12 & 14 \\
 & 3 & 9.60 & 8.8 & 7 & 8 \\
Acetylene & 1 & 9.05 & 60.0\textsuperscript{*} & 21 & 17 \\
 & 2 & 9.60 & 5.0 & 14 & 15 \\
 & 3 & 8.73 & 36.3 & 11 & 11 \\
M-MOF-74 & 1 & 8.58 & 2.5 & 25 & 25 \\
 & 2 & 9.60 & 2.9 & 22 & 22 \\
 & 3 & 8.50 & 1.2 & 9 & 10 \\
Zeolites & 1 & 8.95 & 4.1 & 23 & 19 \\
 & 2 & 8.68 & 9.0 & 13 & 13 \\
 & 3 & 8.70 & 18.9 & 22 & 12 \\
\hline
\end{tabular}
\par\smallskip
\begin{minipage}{\linewidth}
\footnotesize\textsuperscript{*}Approximate managed campaign duration, rather than an autonomous makespan for all requested work. The agent repeatedly requested batches of 20--30 nominally identical, high-cycle simulations per condition. To limit redundant resource use, only one representative from each batch was executed.
\end{minipage}
\end{table}
\FloatBarrier

Across twelve multi-agent runs, the average workflow, simulation-validity, and analysis-consistency scores were 9.36, 8.92, and 9.10, respectively, giving an overall average of 9.13. Deductions concerned normalization errors, limited convergence evidence, or interpretation. Some affected outputs supported only qualitative conclusions or could not support quantitative comparisons. Scores aggregate workflow execution, validity checks, and interpretation. They are not percentages of quantitatively reliable predictions.

The four single-agent baselines scored 6.78--9.05. The comparison is descriptive: each benchmark has one baseline, using GPT-5.4 throughout, and architectures were not matched for model mix, tokens, or inference cost.

\FloatBarrier

\subsection{Force-Field Extraction}
Eight literature-derived cases were evaluated using parameter-set intersection over union (IoU) and RASPA readiness, the fraction of generated force-field files executable without correction. The metrics were combined into $S_{\mathrm{FF}}$ and averaged over repeated extractions (methods in the SI). Table~\ref{tab:ff_results} summarizes performance.

\begin{table}[htbp]
\caption{Force-field extraction performance and weighted score by source.}
\label{tab:ff_results}
\centering
\begin{tabularx}{\linewidth}{Xcccc}
\hline
Method & IoU & Readiness & $S_{\mathrm{FF}}$ & Cost (\$) \\
\hline
CO$_2$ EPM2 \cite{harris1995carbon} & 0.95 & 100\% & 0.970 & 0.19 \\
CO$_2$ in zeolites \cite{garcia2009transferable} & 1.00 & 88\% & 0.952 & 0.23 \\
CO$_2$ in zeolites \cite{fang2013first} & 0.99 & 92\% & 0.962 & 0.34 \\
CO$_2$, N$_2$, O$_2$, Ar in zeolites \cite{vujic2016transferable} & 0.98 & 100\% & 0.988 & 0.24 \\
TraPPE-zeo \cite{bai2013trappe}  & 1.00 & 100\% & 1.000 & 0.15 \\
C$_x$H$_x$, CO$_2$, N$_2$, O$_2$ in zeolites \cite{findley2021transferable} & 1.00 & 100\% & 1.000 & 0.65 \\
TAMOF-1 \cite{gooijer2025tamof} & 0.94 & 75\% & 0.864 & 0.27 \\
UFF + DREIDING (MOFs) \cite{rappe1992uff,mayo1990dreiding} & 1.00 & 100\% & 1.000 & 1.55 \\
\hline
\end{tabularx}
\end{table}

The weighting gives a mean extraction score of 0.967 across eight cases (range 0.864--1.000). Mean parameter IoU is 0.983, indicating close agreement with reference sets. No hallucinated parameters were observed. Deviations mainly reflect omissions during extraction or final-file writing. In a few cases, such as EPM2, irregular source formatting caused incorrect parameter associations. Most issues would be detected by initial simulation logs warning of missing atom or interaction definitions.

Lower readiness arose from incorrect declared atom or interaction counts despite accurate interaction definitions. These bookkeeping errors make the files structurally invalid but recoverable by post-processing. Force-field extraction costs remain low (Table~\ref{tab:ff_results}), increasing when parameters span multiple sources or many entries.

\FloatBarrier

\subsection{Case Study 1: C$_2$H$_2$/CO$_2$ and C$_2$H$_2$/CH$_4$ Separation in Open-Metal-Site MOFs}

\begin{figure}[!htbp]
\centering
\includegraphics[width=0.8\linewidth]{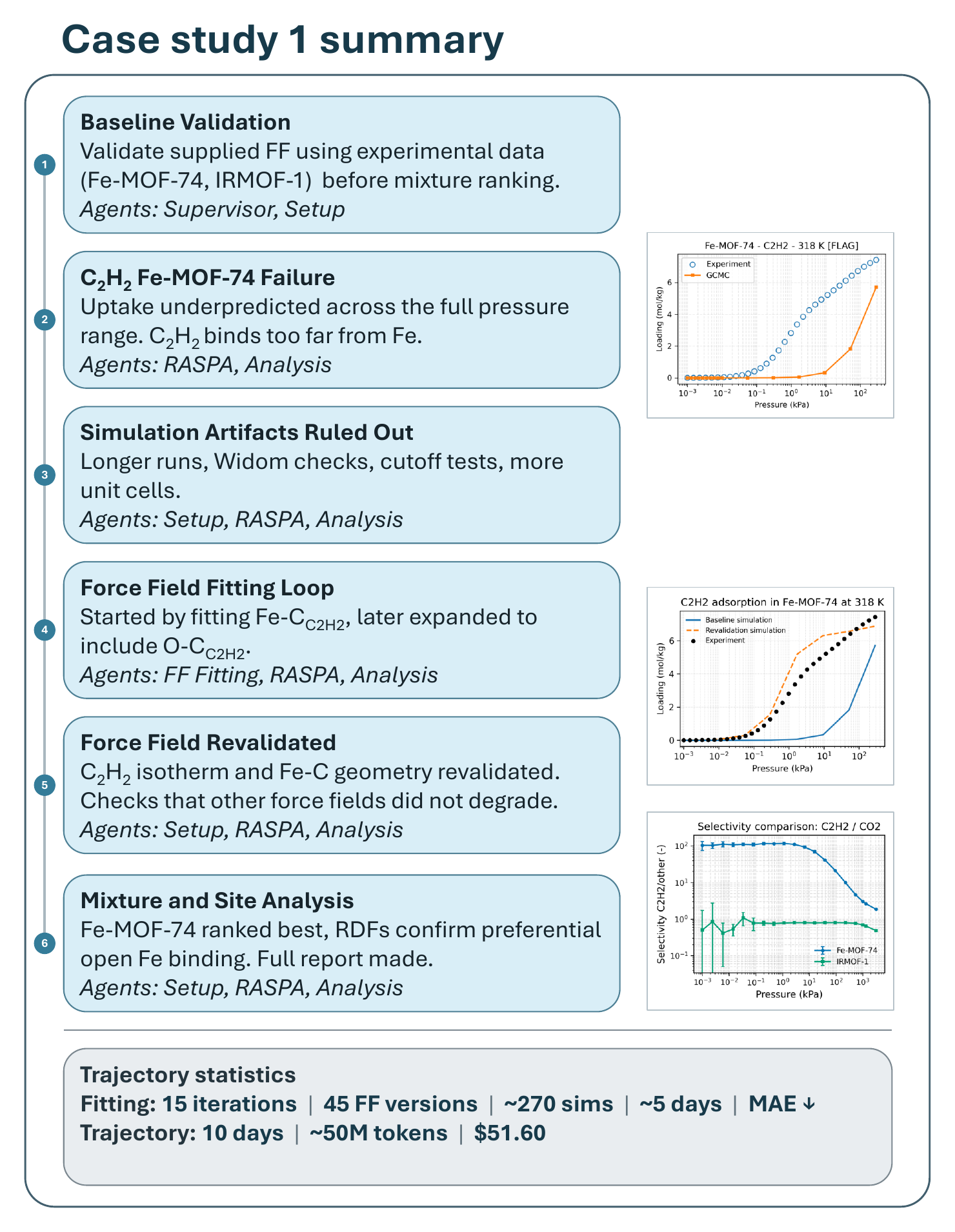}
\caption{Schematic overview of Case Study 1. Here, baseline validation means validation of the supplied initial force field. The workflow includes detection of the poor C$_2$H$_2$ isotherm in Fe-MOF-74, convergence checks, force field fitting, validation of the fitted model, and subsequent mixture adsorption and adsorption site analysis.}
\label{fig:case1_trajectory}
\end{figure}

The first case study tested whether RaspAssistant could identify and correct force-field limitations before analysing acetylene separation in MOFs (Figure~\ref{fig:case1_trajectory}). Initial validation against experimental isotherms revealed substantial disagreement for C$_2$H$_2$ in Fe-MOF-74, while other adsorption results were sufficiently consistent to proceed. The discrepancy persisted after increasing the unit-cell size and Monte Carlo cycles, so RaspAssistant focused on refining the interactions between acetylene and the open Fe sites.

\begin{figure}[!htbp]
\centering
\includegraphics[width=0.6\linewidth]{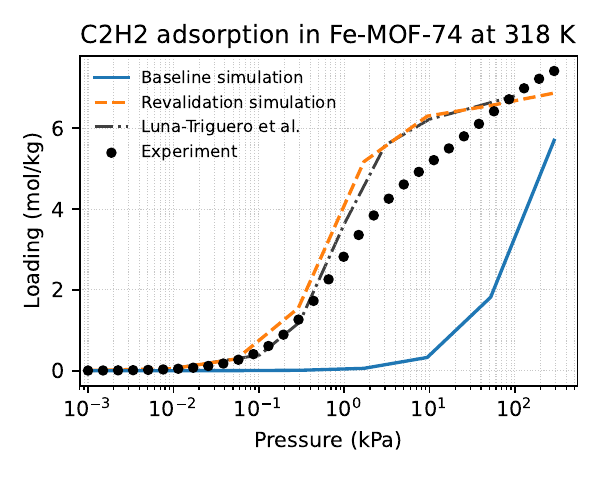}
\caption{C$_2$H$_2$ adsorption isotherms in Fe-MOF-74. ``Baseline simulation'' uses the initial force field. ``Revalidation simulation'' uses the fitted force field. The plot was generated by RaspAssistant. The isotherm from Ref.~\cite{luna2019acetylene} was added manually for comparison.}
\label{fig:fitted}
\end{figure}

Using the fitting harness, RaspAssistant evaluated 45 force-field versions over 15 iterations, with six simulations per version (270 simulations), in approximately 5 days. Initially, the system attempted to correct the force field by fitting only the Fe--C$_{\mathrm{C_2H_2}}$ cross interaction, using the experimental isotherm and the C$_2$H$_2$ binding geometry as fitting targets. This produced partial improvement, but was not sufficient to reproduce the adsorption behaviour. The system subsequently expanded the fitting space to include framework oxygen interactions with the acetylene carbon atom, and later introduced separate interactions for different oxygen environments. This led to a final force field that reproduced the C$_2$H$_2$ adsorption isotherm with accuracy comparable to the force field reported in the reference study, as shown in Figure~\ref{fig:fitted}.

\FloatBarrier

The C$_2$H$_2$ isotherm obtained with the force field fitted by RaspAssistant closely follows the literature-force-field curve over the investigated range. To further assess whether the fitted model captures the local interaction with the open metal site, we compare the predicted C$_2$H$_2$ binding geometry. Force-field fitting reduced the arithmetic mean of the two Fe--C contacts from 3.12 to 2.63~\AA, bringing it close to the literature-force-field value of 2.60~\AA\ and the experimental mean of 2.50~\AA\ (Table~\ref{tab:binding_geometry}).

\begin{table}[htbp]
\centering
\caption{Comparison of the arithmetic mean of the two Fe--C contacts for C$_2$H$_2$ adsorbed in Fe-MOF-74.}
\label{tab:binding_geometry}
\begin{tabular}{lc}
\hline
\textbf{Model} & \textbf{Mean Fe--C distance (\AA)} \\
\hline
Initial force field & 3.12 \\
RaspAssistant fitted force field & 2.63 \\
Literature force field & 2.60 \\
Experiment & 2.50 \\
\hline
\end{tabular}
\end{table}

The final fitted parameters differ substantially from the literature model (Table~\ref{tab:ff_comparison}), despite similar isotherms: RaspAssistant concentrates the correction in a stronger Fe--C$_{\mathrm{C_2H_2}}$ interaction with weaker oxygen contributions, whereas the literature correction is distributed over several framework atom types.

\begin{table}[htbp]
\centering
\caption{Comparison of fitted Lennard--Jones parameters for acetylene--framework interactions in Fe-MOF-74. Literature parameters are taken from Ref.~\cite{luna2019acetylene}.}
\label{tab:ff_comparison} \begin{tabular}{llcccc} \hline & & \multicolumn{2}{c}{Agent} & \multicolumn{2}{c}{Literature} \\ \cline{3-6} Interaction Pair & Atom Types & $\epsilon$ (K) & $\sigma$ (\AA) & $\epsilon$ (K) & $\sigma$ (\AA) \\ \hline Metal site & Fe -- C$_{\mathrm{C_2H_2}}$ & 680.0 & 2.355 & 38.92 & 2.878 \\ \hline Framework oxygen & O1 -- C$_{\mathrm{C_2H_2}}$ & 10.0 & 2.900 & 79.23 & 2.870 \\ & O2 -- C$_{\mathrm{C_2H_2}}$ & 20.0 & 3.050 & 79.23 & 2.870 \\ & O3 -- C$_{\mathrm{C_2H_2}}$ & 20.0 & 3.050 & 79.23 & 2.870 \\ \hline Framework carbon & C1/2/3/4 -- C$_{\mathrm{C_2H_2}}$ & --- & --- & 78.96 & 3.055 \\ \hline Framework hydrogen & H -- C$_{\mathrm{C_2H_2}}$ & --- & --- & 31.57 & 2.792 \\ \hline \end{tabular}
\par\smallskip
\begin{minipage}{\linewidth}
\footnotesize Dashes indicate no fitted cross-interaction override. The initial force field and mixing rules remain active for these pairs.
\end{minipage}
\end{table}

Variation in the auxiliary CO$_2$ guardrail minimizations may have caused the fitting loop to reject candidates that were otherwise moving the C$_2$H$_2$ description in a promising direction.

The RaspAssistant optimization used the adsorption isotherm and binding geometry as fitting targets, whereas the literature model was also validated against adsorption-heat data \cite{luna2019acetylene}. Experimental adsorption heats were unavailable during fitting. The subsequent $Q_{\mathrm{st}}$ comparison therefore provides an independent energetic assessment. The fitted model captures the low-loading energy scale and the qualitative decrease toward saturation, although the decrease is sharper than in experiment. The literature model follows the low-loading data more closely but also deviates over part of the high-loading region (Figure~\ref{SI-fig:si_acetylene_qst}). This broad agreement outside the fitting targets is encouraging, while showing how adsorption-heat data could further constrain future fitting campaigns.

\begin{figure}[!htbp]
\centering
\includegraphics[width=\linewidth]{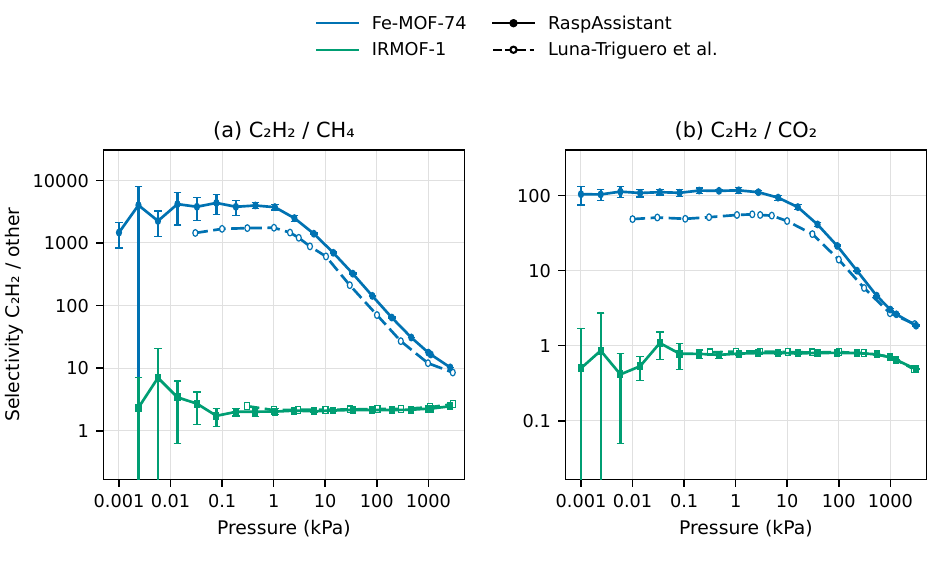}
\caption{Comparison of C$_2$H$_2$/CH$_4$ (a) and C$_2$H$_2$/CO$_2$ (b) mixture selectivities from the fitted Fe-MOF-74 and retained IRMOF-1 models with Ref.~\cite{luna2019acetylene}. RaspAssistant generated the plot. Reference data were added manually.}
\label{fig:case1_selectivity}
\end{figure}

After fitting, RaspAssistant simulated mixtures and calculated C$_2$H$_2$/CO$_2$ and C$_2$H$_2$/CH$_4$ selectivities. These generally agree with literature trends for Fe-MOF-74 and IRMOF-1 (Figure~\ref{fig:case1_selectivity}). Deviations are largest when the less strongly adsorbed component has very low uptake, making selectivity sensitive to small absolute loading differences.

\begin{figure}[!htbp]
\centering
\includegraphics[width=0.75\linewidth]{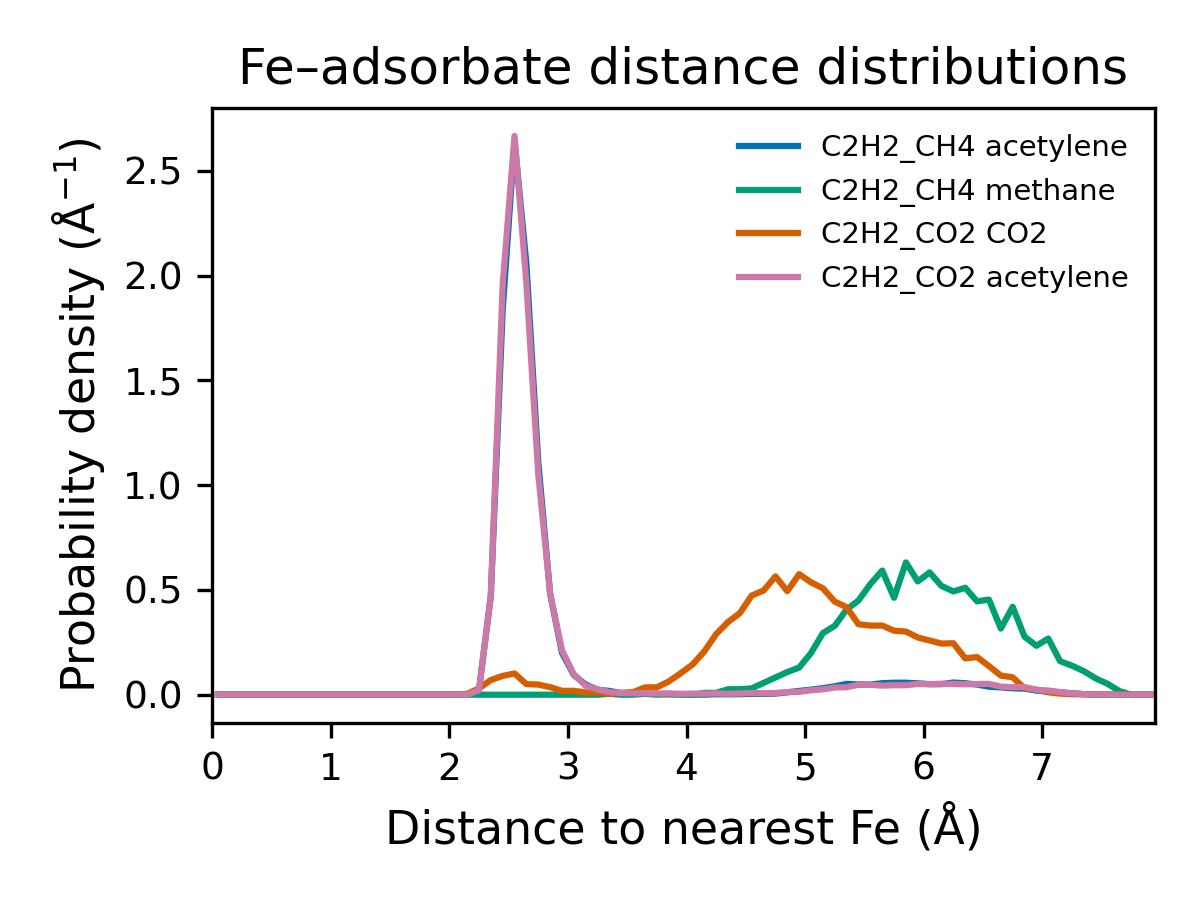}
\caption{Normalized minimum periodic interaction-site-to-Fe distance distributions in Fe-MOF-74 for equimolar C$_2$H$_2$/CH$_4$ and C$_2$H$_2$/CO$_2$ mixtures at 298~K and 1000~kPa. The sharp short distance peak for C$_2$H$_2$ indicates preferential localization near the open Fe sites, whereas CO$_2$ and CH$_4$ are mainly distributed farther away in the pore space.}
\label{fig:case1_rdf}
\end{figure}

RaspAssistant then interpreted the adsorption sites through nearest Fe distance distributions (Figure~\ref{fig:case1_rdf}). C$_2$H$_2$ is strongly localized near the open Fe sites, whereas CO$_2$ and CH$_4$ have broader distributions farther into the pore space, consistent with the physical picture from the reference study's spatial histograms.

Case Study 1 required approximately 10 wall-clock days, 50 million tokens, and US\$51.6 in API-equivalent cost.

\FloatBarrier

\subsection{Case Study 2: High-Throughput Screening of Pure-Silica Zeolites for CO$_2$ Adsorption}

\begin{figure}[!htbp]
\centering
\includegraphics[width=0.8\linewidth]{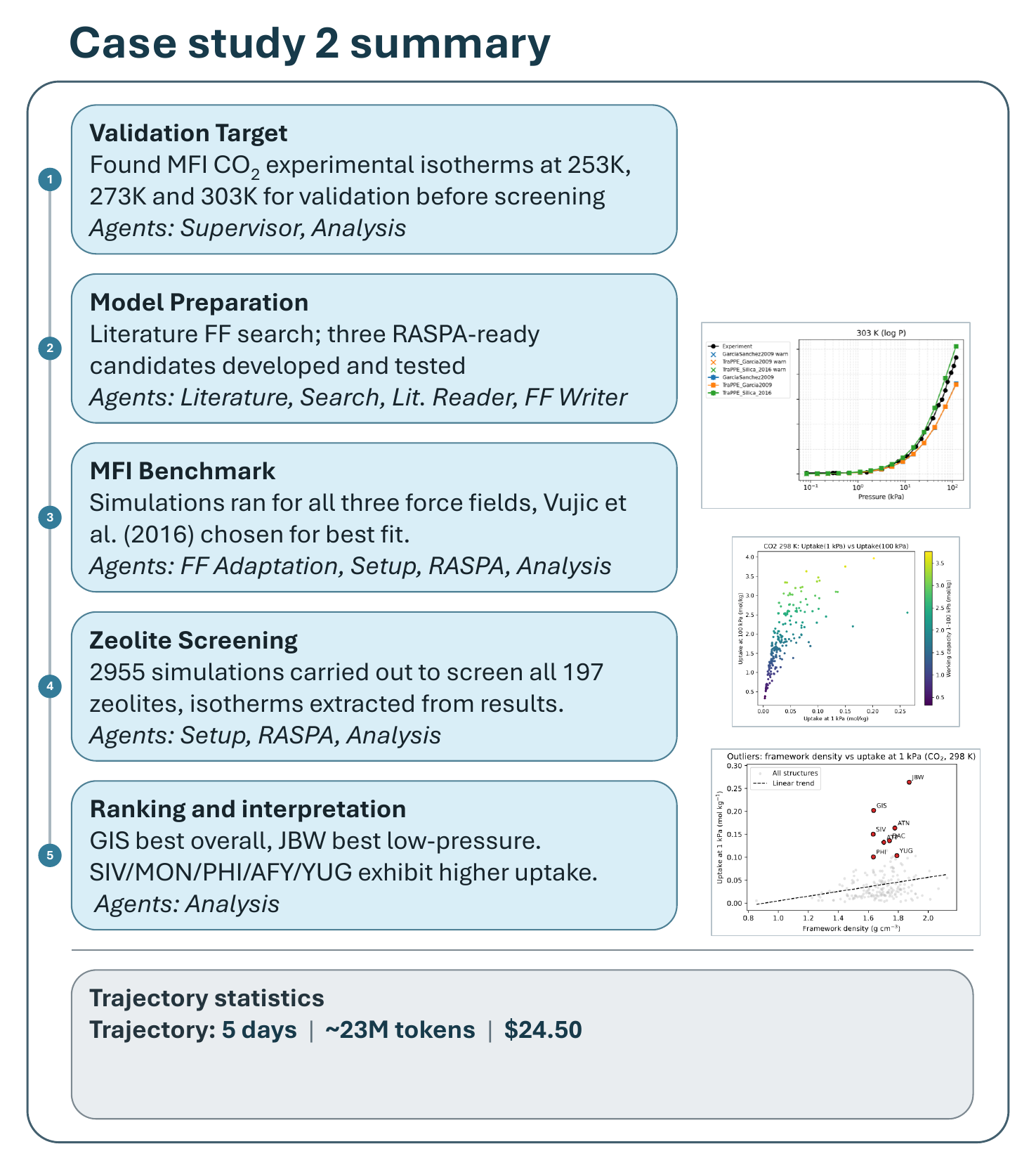}
\caption{Schematic overview of the main steps taken by RaspAssistant in Case Study 2. The workflow includes preparation of literature-based force-field candidates, validation against MFI CO$_2$ adsorption experiments, selection of the best-performing force field, 298 K screening of the pure-silica zeolite dataset, and post-processing through ranking and descriptor analysis.}
\label{fig:case2_trajectory}
\end{figure}

We next tested whether experimental validation could guide force-field selection before applying a common screening protocol across pure-silica zeolites (Figure~\ref{fig:case2_trajectory}). Given MFI experimental isotherms and pure-silica zeolite structures, RaspAssistant prepared three CO$_2$ force fields: the Garcia-Sanchez model \cite{garcia2009transferable}, its framework model combined with TraPPE CO$_2$ \cite{garcia2009transferable,potoff2001vapor}, and TraPPE CO$_2$ combined with a Vujic-derived silica-zeolite framework model \cite{vujic2016transferable}. Simulated MFI isotherms were compared with experiment at 253, 273, and 303 K.

\FloatBarrier

\begin{figure}[!htbp]
\centering
\includegraphics[width=\linewidth]{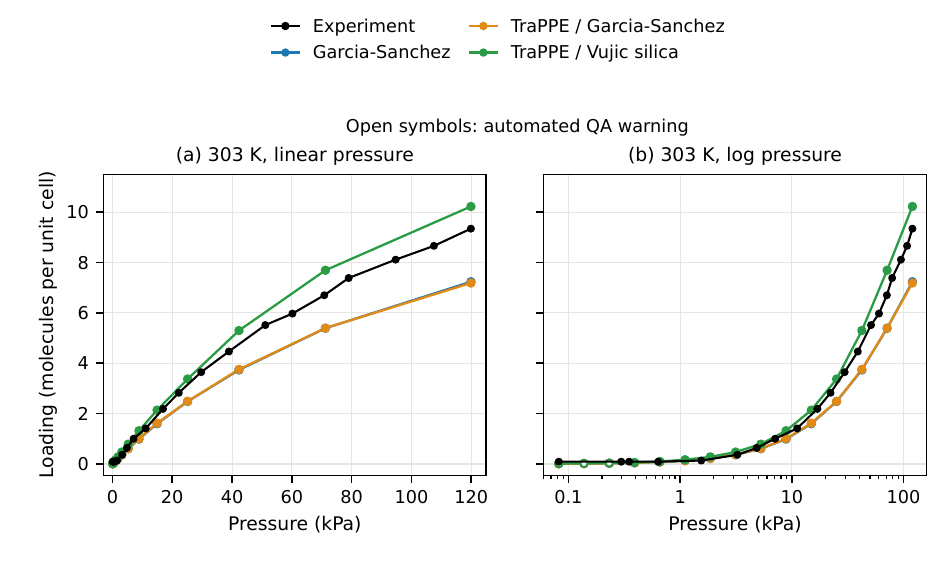}
\caption{Validation of the candidate force fields against the provided CO$_2$ adsorption isotherm in MFI at 303 K. The \texttt{TraPPE\_Silica\_2016} candidate combines TraPPE CO$_2$ with a Vujic-derived silica-zeolite model and gives the best agreement. Open markers flag block-drift outliers. These points are retained in the error calculations (see the SI).}
\label{fig:case2_mfi_validation}
\end{figure}

The TraPPE CO$_2$/Vujic-derived silica model best matched MFI experiments overall, with a weighted mean absolute error (MAE) of 0.72 molecules per unit cell, compared with 0.866 for Garcia-Sanchez and 0.872 for Garcia-Sanchez/TraPPE-Potoff. At 303 K (Figure~\ref{fig:case2_mfi_validation}), the Vujic-derived candidate~\cite{vujic2016transferable} follows experiment more closely over most of the pressure range. The other models underpredict more strongly, while the selected model slightly overpredicts at higher pressures. RaspAssistant selected it for screening, treating results as relative rankings under a consistent protocol rather than quantitatively exact predictions.

\FloatBarrier

\begin{figure}[!htbp]
\centering
\includegraphics[width=\linewidth]{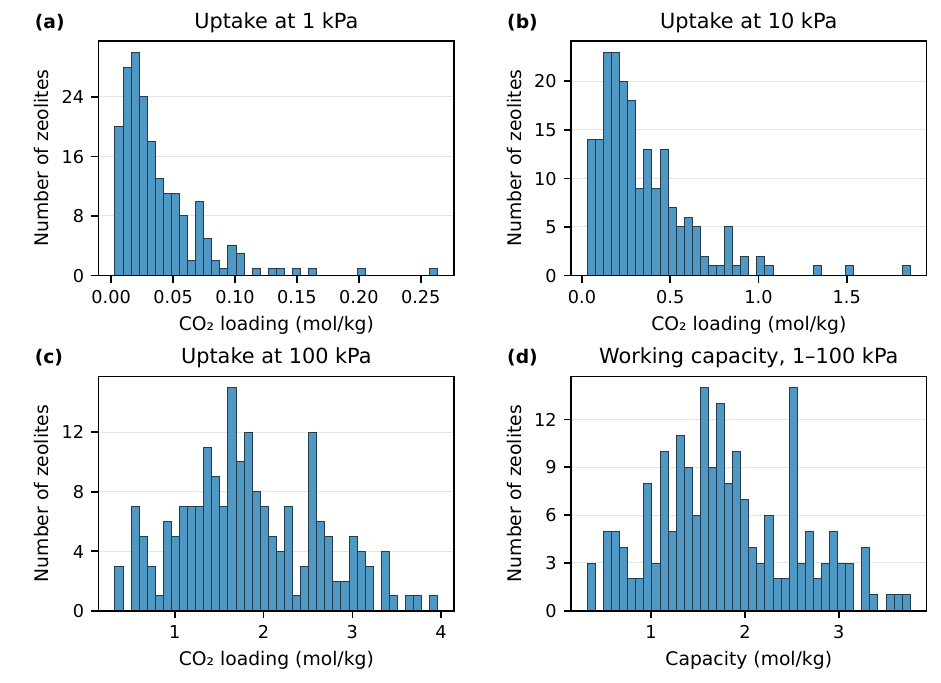}
\caption{Distribution of interpolated CO$_2$ uptake at 1, 10, and 100 kPa and of the derived 1--100 kPa working capacity in Case Study 2.}
\label{fig:case2_uptake_p}
\end{figure}

RaspAssistant completed 2,955 CO$_2$ adsorption calculations at 298 K: 15 pressures for each of 197 structures passing automated input checks. Outputs were checked for missing or nonphysical values affecting ranking. Isotherms provided interpolated uptake at 1, 10, and 100 kPa and working capacity $q(100~\mathrm{kPa})-q(1~\mathrm{kPa})$ (Figure~\ref{fig:case2_uptake_p}). Rankings changed modestly with pressure: JBW led at 1 kPa, while GIS led at 10 and 100 kPa and in working capacity. SIV, MON, PHI, AFY, and YUG were also among the strongest candidates at higher pressure or for working capacity.

\FloatBarrier

\begin{figure}[!htbp]
\centering
\includegraphics[width=\linewidth]{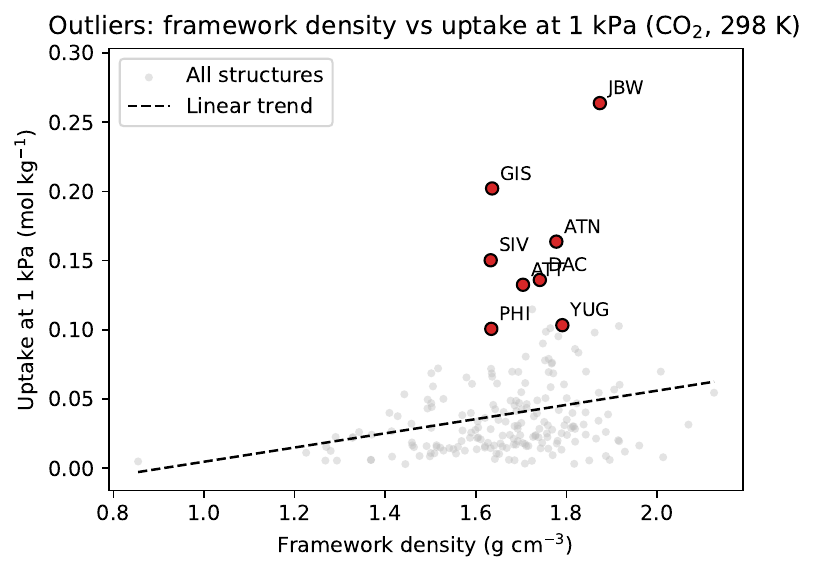}
\caption{Relationship between framework density and simulated CO$_2$ uptake at 1 kPa for the screened pure-silica zeolites. Highlighted topologies correspond to positive outliers relative to the overall linear trend, indicating unusually high uptake for their framework density.}
\label{fig:case2_density_outliers}
\end{figure}

Framework density showed a weak positive association with simulated CO$_2$ uptake at 1 kPa (Spearman $\rho=0.284$, Figure~\ref{fig:case2_density_outliers}). GIS, JBW, SIV, ATN, and YUG were positive residual outliers. The retained descriptors lacked pore size, accessible volume, and channel connectivity, so these residuals cannot be assigned to a specific pore-scale mechanism.

Case Study 2 required approximately 5 wall-clock days, 23 million tokens, and US\$24.5 in API-equivalent cost.

\FloatBarrier

\section{Discussion}
RaspAssistant connects literature-derived interaction models, experimental validation, atomistic simulation, and separation analysis within a coordinated research workflow. Its contribution extends beyond preparing or executing individual simulations to deciding how a study should proceed as results become available. The open-ended studies demonstrate this capability through refinement of an acetylene force field and validation-guided selection of a CO$_2$ model before zeolite screening. These decisions connect the adequacy of the molecular description to the interpretation of downstream predictions.

A consequential design choice is the combination of high-level scientific planning with direct access to complete simulation and force-field files. The Supervisor coordinates specialist agents while retaining the research objective, and the specialists can revise inputs, perform calculations, and return analyses that inform subsequent planning. This gives the framework flexibility to change both model details and the sequence of calculations within a study. In the acetylene case, the fitting procedure expanded the interaction model in response to the observed adsorption behaviour, then returned the fitted model to mixture and site analysis. The fitting loop itself follows a controlled sequence of parameter selection, simulation, analysis, and continuation decisions. Flexible study planning and structured model optimization therefore serve complementary roles.

The evaluations test different parts of this research process. Published-workflow reproduction examines coordination across established adsorption and separation methods, while literature-based extraction examines reconstruction of the interaction models needed for simulation. The open-ended cases then test whether intermediate results can guide the choice of subsequent calculations. Using the same framework across these tasks supports a broader role for agentic tools in connecting model construction, validation, and application. The descriptive architectural baseline does not isolate which individual design choices account for the observed performance.

The scientific value of this integration is clearest when several types of evidence are considered together. In acetylene adsorption, similar isotherms from substantially different fitted and literature parameter sets show why a single fitting target cannot fully establish model adequacy. Experimental heats unavailable during fitting provided an independent energetic assessment, complementing the isotherm, binding geometry, mixture, and site analyses. In zeolite screening, validation in MFI guided model selection before applying a common protocol across the structure set. Pressure-dependent rankings then connected candidate selection to operating conditions. Such rankings identify candidates for further investigation, with quantitative transferability beyond the reference system remaining a separate question.

Further development can build on this integration by connecting screening more closely to physical interpretation. Pore geometry descriptors from tools such as Zeo++ \cite{willems2012algorithms} could relate adsorption trends to accessible volume and channel structure, while systematic uncertainty analysis could connect confidence in molecular predictions to subsequent mixture and process comparisons. Shared workflow records also provide a basis for expert review of how model choices and intermediate results lead to a reported conclusion. The resulting division of work allows agents to coordinate and document extended computational studies after task specification and input provision, with researchers defining the objectives and assessing the physical models and scientific conclusions.

\section{Methods}
RaspAssistant receives a natural-language objective and a folder of structures, experimental data, and available force fields. Each benchmark run began without benchmark-specific memory from earlier runs. Computational methods and agent configurations are in the SI.

A Supervisor agent decomposes the research objective and coordinates specialists for literature retrieval, force-field preparation, simulation setup and execution, and analysis. Higher-level agents plan, execute, verify, and revise their subtasks. Specialists can read and modify complete files, execute calculations, and analyse outputs. Molecular simulations used RASPA 2.0.47 through a local simulation interface, with RUPTURA supporting mixture and breakthrough calculations. RASPA outputs are converted to structured JSON for analysis, and shared per-run memory and stored scientific files retain context across delegated tasks. The force-field fitting controller organizes successive parameter-selection, simulation, and analysis steps. Agents made scientific decisions during workflow execution. Benchmark execution procedures and human involvement are detailed in the SI.

One human assessor scored workflow-reproduction runs for workflow correctness, simulation validity, and analysis consistency with weights of 40\%, 40\%, and 20\%. Force-field extraction combines exact parameter-set agreement (60\%) and direct RASPA readiness (40\%). Scoring prioritized completed simulation outputs and effective settings over final reports. Downstream results were assessed in light of upstream validity. Full methods, configurations, benchmark definitions, and scoring procedures are in the SI.

\section*{Data availability}

The supporting data are available on Zenodo at \href{https://doi.org/10.5281/zenodo.23462699}{\nolinkurl{10.5281/zenodo.23462699}}. The package contains chronological JSON and HTML workflow records, final reports and evaluation records, source data for figures and tables, and case-study force-field definitions and numerical summaries. An artifact manifest links the records to the scientific files.

\section*{Code availability}

The RaspAssistant source code and production prompt library are not publicly released. The frozen source code, prompts, and configuration used for the experiments reported here are available confidentially to editors and reviewers upon request. Manuscript-specific analysis files and source data are included in the supporting data package described in the Data Availability statement.

\section*{Supporting Information}
Computational methods and agent configurations, detailed benchmark results, workflow-reproduction prompts and evaluation criteria, supporting workflow diagrams and adsorption analyses, and force-field fitting procedure and independent energetic validation (PDF).

\par\medskip
\noindent\begin{minipage}{\linewidth}
\section*{Author contributions}
M.P. conceived the study, developed the framework, carried out the simulations and analyses, and drafted the manuscript. S.G. and J.M.V.L. contributed to the design of input files for the different use cases. V.M. and S.C. supervised the research. Funding was obtained by S.C. and M.P. All authors reviewed the manuscript.
\end{minipage}
\par

\section*{Competing interests}
The authors declare no competing interests.

\section*{Acknowledgment}
OpenAI Codex was used to assist with editing the manuscript in October 2026.

\providecommand{\latin}[1]{#1}
\makeatletter
\providecommand{\doi}
  {\begingroup\let\do\@makeother\dospecials
  \catcode`\{=1 \catcode`\}=2 \doi@aux}
\providecommand{\doi@aux}[1]{\endgroup\texttt{#1}}
\makeatother
\providecommand*\mcitethebibliography{\thebibliography}
\csname @ifundefined\endcsname{endmcitethebibliography}
  {\let\endmcitethebibliography\endthebibliography}{}

\clearpage
\setcounter{section}{0}
\setcounter{subsection}{0}
\setcounter{figure}{0}
\setcounter{table}{0}
\setcounter{equation}{0}
\setcounter{footnote}{0}
\renewcommand{\thefigure}{S\arabic{figure}}
\renewcommand{\thetable}{S\arabic{table}}
\renewcommand{\theequation}{S\arabic{equation}}
\renewcommand{\theHsection}{SI.\arabic{section}}
\renewcommand{\theHfigure}{SI.\arabic{figure}}
\renewcommand{\theHtable}{SI.\arabic{table}}
\renewcommand{\theHequation}{SI.\arabic{equation}}
\makeatletter
\global\let\maketitle\PreprintMakeTitle
\global\let\@maketitle\PreprintInnerTitle
\global\let\thanks\PreprintThanks
\global\let\@author\PreprintAuthor
\gdef\@title{Supporting Information for\\[0.5em]
\textit{RaspAssistant as an Agentic Framework for Automating Adsorption Simulation Research in Porous Materials}}
\gdef\@date{}
\gdef\@thanks{}
\makeatother

\maketitle

\noindent\textbf{Reader guide.} Expanded research context is in Section~\ref{SI-sec:additional_context}. Run-level benchmark results are in Section~\ref{SI-sec:workflow_results}. Section~\ref{SI-sec:computational_methods} describes the framework. Detailed scoring, original task prompts, and scientific criteria are in Sections~\ref{SI-sec:workflow_evaluation}, \ref{SI-sec:task_prompts}, and \ref{SI-sec:benchmark_criteria}. Fitting and energetic validation are in Sections~\ref{SI-sec:fitting_harness} and \ref{SI-sec:energetic_validation}. Prompt headings identify the study number used in the results.

\section{Additional Research Context}
\label{SI-sec:additional_context}

\subsection{Additional Context on Agentic Scientific Systems}

Recent advances in artificial intelligence (AI) have led to the emergence of agentic scientific systems, in which large language models (LLMs) are combined with planning, memory, tool use, and iterative reasoning to execute complex scientific tasks \cite{SI-dral2026machine}. In experimental science, ChemCrow \cite{SI-bran2024chemcrow} and Coscientist \cite{SI-boiko2023coscientist} demonstrate how LLM-based agents can integrate literature knowledge, planning, and tool use, while autonomous materials laboratories \cite{SI-burger2020mobilechemist,SI-abolhasani2023sdl,SI-szymanski2023alab} illustrate the broader closed-loop integration of computation and experiment. These developments suggest a broader shift from AI systems that assist with isolated subtasks toward systems that can coordinate multi-step scientific workflows.

A similar trend is emerging in computational chemistry and materials simulation. The El Agente family \cite{SI-zou2025agente,SI-gustin2025agente,SI-choi2026agente,SI-perez2026agente} has shown that multi-agent architectures can translate natural-language instructions into executable quantum-chemistry, molecular-editing, and quantum-simulation workflows. Systems including DREAMS \cite{SI-wang2025dreams}, VASPilot \cite{SI-liu2025vaspilot}, MASTER \cite{SI-rothfarb2026master}, MatSciAgent \cite{SI-chaudhari2026matsciagent}, Masgent \cite{SI-liu2026masgent}, and an end-to-end multi-agent framework for LAMMPS workflows \cite{SI-vriza2026atomistic} extend this approach to tasks such as simulation setup, convergence testing, job submission, error recovery, structure generation, data retrieval, and molecular dynamics. More recent systems support autonomous interatomic-potential selection and validation through URSA \cite{SI-somasundaram2026ursa}, traceable multi-method materials workflows through ALKEMIE Agent \cite{SI-huang2026alkemie}, and adaptive planning and execution across several simulation methods through QUASAR \cite{SI-yang2026quasar}. AtomAgents combines multi-agent reasoning with physics-based alloy design \cite{SI-ghafarollahi2025atomagents}, while MDAgent generates, executes, and iteratively reviews LAMMPS scripts for thermodynamic calculations \cite{SI-shi2025mdagent}. These implementations expose simulation control at different levels. For example, ChemGraph \cite{SI-pham2026chemgraph} coordinates computational chemistry through structured interfaces to tools such as the Atomic Simulation Environment (ASE), whereas GENIUS \cite{SI-soleymanibrojeni2026genius} generates and revises Quantum ESPRESSO input files within a knowledge-graph-guided finite-state workflow. Benchmarking of LLM-generated LAMMPS inputs \cite{SI-holbrook2026lammps} further illustrates the challenge of combining low-level control with scientifically valid setup. Thus, flexibility must be assessed with respect to the decisions exposed by a particular interface and workflow rather than treated as a binary property of an agentic framework.

Within porous materials, domain-specific agentic automation is developing rapidly, as surveyed by Yu and co-workers \cite{SI-yu2026mofagents}. SimMOF \cite{SI-lee2026simmof} connects natural-language queries to Zeo++ \cite{SI-willems2012algorithms}, RASPA \cite{SI-dubbeldam2016raspa}, VASP \cite{SI-kresse1993ab}, and LAMMPS \cite{SI-thompson2022lammps}, and demonstrates geometric analysis, adsorption, diffusion, binding-energy analysis, and methane-uptake screening. SimMOF generates simulation inputs using LLM reasoning and retrieved literature, and can revise inputs and rerun calculations during error recovery \cite{SI-lee2026simmof}. Our previous multi-agent framework \cite{SI-petkovic2025fullyautomated} demonstrated literature-informed force-field extraction and automated RASPA setup. Other recent MOF studies combine multi-agent evaluation with molecular simulation and process modelling for CO$_2$-capture screening \cite{SI-li2026mofscreening}, or literature-derived rules, GCMC, and machine learning in a human-supervised radon-capture workflow \cite{SI-zhang2026radonagent}. Agent-assisted sorption campaigns have also combined deterministic GCMC--MD execution with selective LLM review in clay--water desorption \cite{SI-wang2026agentmd}. These studies demonstrate connected simulation workflows, input preparation, and screening. Among the systems discussed here, we found no reported evaluation reproducing complete published adsorption and separation studies across the combined sequence of force-field selection or adaptation, experimental validation, pure-component and mixture simulations, and downstream IAST or breakthrough analysis.

\section{Detailed Workflow-Reproduction Results}
\label{SI-sec:workflow_results}

\subsection{Study 1: CO$_2$/N$_2$ Separation in TAMOF-1.}
\begin{table}[htb]
\centering
\caption{Multi-agent RaspAssistant performance and the single-agent baseline on Study 1.}
\label{SI-tab:tamof1}
\begin{tabular}{l c c c c}
\hline
\textbf{Run} & \textbf{Workflow} & \textbf{Simulations} & \textbf{Analysis} & \textbf{Overall} \\
\hline
Run 1 & 10.00 & 10.00 & 10.00 & 10.00 \\
Run 2 & 10.00 & 9.00 & 10.00 & 9.60 \\
Run 3 & 10.00 & 9.25 & 9.50 & 9.60 \\
\hline
\textbf{Multi-agent average} & \textbf{10.00} & \textbf{9.42} & \textbf{9.83} & \textbf{9.73} \\
\hline
Single-agent baseline & 9.13 & 9.13 & 8.75 & 9.05 \\
\hline
\end{tabular}
\end{table}
\FloatBarrier

As summarized in Table~\ref{SI-tab:tamof1}, the three multi-agent runs completed the requested pure-component, mixture, ideal adsorbed solution theory (IAST), and breakthrough workflow. Run~1 received full credit. Runs~2 and 3 lost points mainly for limited formal convergence evidence. Run~2 also used a slightly reduced cutoff, and Run~3 made one overgeneralized temperature-trend statement. Breakthrough disagreement with experiment was not penalized. The single-agent baseline scored 9.05: its final pure, mixture, and breakthrough outputs were complete, with deductions for the absent IAST--GCMC check, partial breakthrough-validation coverage, provisional breakthrough fits, and limited qualification of selectivities when N$_2$ loading was small.

\subsection{Study 2: C$_2$H$_2$/CO$_2$ and C$_2$H$_2$/CH$_4$ Separation in Open-Metal-Site MOFs.}
\begin{table}[htb]
\centering
\caption{Multi-agent RaspAssistant performance and the single-agent baseline on Study 2.}
\label{SI-tab:acetylene}
\begin{tabular}{l c c c c}
\hline
\textbf{Run} & \textbf{Workflow} & \textbf{Simulations} & \textbf{Analysis} & \textbf{Overall} \\
\hline
Run 1 & 9.50 & 8.13 & 10.00 & 9.05 \\
Run 2 & 10.00 & 9.00 & 10.00 & 9.60 \\
Run 3 & 8.88 & 8.56 & 8.75 & 8.73 \\
\hline
\textbf{Multi-agent average} & \textbf{9.46} & \textbf{8.56} & \textbf{9.58} & \textbf{9.13} \\
\hline
Single-agent baseline & 7.50 & 8.56 & 8.13 & 8.05 \\
\hline
\end{tabular}
\end{table}
\FloatBarrier

As summarized in Table~\ref{SI-tab:acetylene}, the three multi-agent runs completed the principal pure-component, mixture, binding-geometry, and adsorption-site workflow. Run~1 lost points for partial heat coverage, unresolved missing-interaction warnings, limited independent convergence evidence, and a small high-pressure fugacity allowance. Its repeated replica requests were treated as a resource-planning opportunity rather than an accuracy failure. Run~2 lost points for limited convergence evidence and the same fugacity allowance. Run~3 transparently excluded Fe-MOF-74 from unsupported mixture predictions after its C$_2$H$_2$ interaction could not be validated, but lost points for sparse Fe-MOF-74 validation and heat coverage, incomplete mixture coverage, limited convergence evidence, and a site-analysis choice based on non-matching pressure endpoints. The negative binding-geometry result and numerical disagreement with experiment were not penalized. The single-agent baseline scored 8.05, with deductions for omitted Fe-MOF-74 result families, unavailable heat validation, and an indirect site analysis.

\subsection{Study 3: Metal Center Effects in M-MOF-74.}
\begin{table}[htb]
\centering
\caption{Multi-agent RaspAssistant performance and the single-agent baseline on Study 3.}
\label{SI-tab:mof74}
\begin{tabular}{l c c c c}
\hline
\textbf{Run} & \textbf{Workflow} & \textbf{Simulations} & \textbf{Analysis} & \textbf{Overall} \\
\hline
Run 1 & 8.69 & 8.75 & 8.00 & 8.58 \\
Run 2 & 9.50 & 9.75 & 9.50 & 9.60 \\
Run 3 & 8.50 & 8.50 & 8.50 & 8.50 \\
\hline
\textbf{Multi-agent average} & \textbf{8.90} & \textbf{9.00} & \textbf{8.67} & \textbf{8.89} \\
\hline
Single-agent baseline & 6.88 & 7.25 & 5.63 & 6.78 \\
\hline
\end{tabular}
\end{table}
\FloatBarrier

As summarized in Table~\ref{SI-tab:mof74}, the three multi-agent runs completed pure-component validation, 298~K fitting, mixture checks, breakthrough, and adsorption-site analysis, and used direct mixture GCMC to qualify unstable IAST predictions. Run~1 lost points mainly because its H$_2$ and heat validation used the wrong metal-site normalization, with smaller deductions for convergence evidence and site interpretation. Run~2 lost points for incomplete H$_2$ heat validation, a small high-pressure assumption allowance, and an overly fine breakthrough ordering among close cases. Run~3 used the wrong H$_2$/heat basis and received smaller deductions for convergence evidence and reliance on qualified breakthrough rankings. These normalization issues did not invalidate independently normalized mixture results. The single-agent baseline scored 6.78 because it lacked a valid 298~K H$_2$ process basis, complete all-feed and breakthrough validation, and full-metal-site geometry analysis.

\subsection{Study 4: Olefin/Paraffin Separation with Zeolites.}
\begin{table}[htb]
\centering
\caption{Multi-agent RaspAssistant performance and the single-agent baseline on Study 4.}
\label{SI-tab:olepara}
\begin{tabular}{l c c c c}
\hline
\textbf{Run} & \textbf{Workflow} & \textbf{Simulations} & \textbf{Analysis} & \textbf{Overall} \\
\hline
Run 1 & 9.50 & 8.50 & 8.75 & 8.95 \\
Run 2 & 9.00 & 8.31 & 8.75 & 8.68 \\
Run 3 & 8.75 & 9.25 & 7.50 & 8.70 \\
\hline
\textbf{Multi-agent average} & \textbf{9.08} & \textbf{8.69} & \textbf{8.33} & \textbf{8.78} \\
\hline
Single-agent baseline & 9.38 & 8.63 & 5.00 & 8.20 \\
\hline
\end{tabular}
\end{table}
\FloatBarrier

As summarized in Table~\ref{SI-tab:olepara}, Run~1 completed all principal calculation families and retained detailed simulation and analysis records. Its main deduction was propagation of unity rather than final Rosenbluth factors for flexible C$_4$--C$_6$ adsorbates: C$_3$ remained valid, the C$_4$/C$_5$ results were retained only for qualitative interpretation, and C$_6$ was not quantitatively interpretable. Smaller deductions reflected limited convergence evidence and incomplete synthesis of available rankings. Run~2 completed the validation, 288-mixture, and five-map grids and correctly applied 298~K Rosenbluth factors to mixtures. It lost points for unity factors in the C$_6$ isobars, an undersized ITQ-29 cell, omitted cation-position restarts, and overprecise small-denominator selectivities. Run~3 completed all requested simulation families, including 832 direct-mixture calculations with appropriate periodic cells and correctly propagated 298~K Rosenbluth factors. Its deductions were localized to the C$_6$ validation series, which used unity rather than the calculated temperature-dependent factors and was not compared directly with experiment, together with limited convergence evidence and overprecise reporting of some small-denominator selectivities. The single-agent baseline scored 8.20. After an autonomous repair completed its grid, deductions remained for the undersized ITQ-29 cell and insufficient qualification of small-denominator and site-mechanism claims.

\section{Computational Methods}
\label{SI-sec:computational_methods}

\subsection{RaspAssistant}

\subsubsection{Overview}

RaspAssistant is a hierarchical multi-agent framework designed to automate porous materials simulations, from literature-informed force field construction to simulation execution and analysis. As illustrated in Figure~\ref{fig:raspassist}, the system integrates two main functional components: (i) a simulation pipeline responsible for setting up, executing, and analyzing molecular simulations, and (ii) a force-field extraction pipeline that retrieves and structures information from scientific publications. These components are tightly coupled, enabling the system to incorporate literature-derived knowledge directly into simulation workflows.

Agents receive a natural-language description of the research objective, together with a structured folder containing relevant data, including experimental measurements, structural information (e.g., CIF files), and force field definitions when available. This design allows the framework to reproduce published workflows as closely as possible, while reducing computational cost and ensuring controlled evaluation. The primary output of the system is a structured report addressing the requested task (e.g., gas separation performance), although evaluation focuses on the full workflow, including planning, simulation validity, and analysis.

\subsubsection{System Architecture}

The framework follows a hierarchical organization, with a Supervisor Agent responsible for high-level scientific planning and coordination. Based on the input task and available resources, the Supervisor Agent decomposes the problem into a sequence of subtasks, determines the required simulations, and delegates responsibilities to specialized agents. It can also invoke a File agent for managing directory structures and organizing intermediate and final results.

The simulation pipeline consists of several agents. The Setup agent generates simulation inputs by combining information from the task description, the RASPA manual, and prior experience stored in memory. It is supported by auxiliary agents, including an Example agent that retrieves the location of relevant simulation examples from the RASPA manual and a Python agent for automated file manipulation and template generation.

A distinction is made between simulation setup and execution depending on the complexity of the underlying software. For RASPA simulations, input generation is non-trivial, and responsibilities are split between a Setup agent, which prepares simulation inputs, and a dedicated RASPA agent, which executes simulations and performs initial diagnostics (e.g., detection of runtime errors or incomplete jobs). In contrast, for RUPTURA simulations, setup and execution are sufficiently lightweight to be handled by a single agent. In addition to running RUPTURA tasks, the RUPTURA agent performs basic validation of results, including inspection of isotherm fits and derived quantities such as IAST predictions and breakthrough curves.

In addition, an Analysis agent performs in-depth interpretation of simulation outputs, including post-processing, consistency checks, and visualization. RASPA outputs can contain thousands of lines of unstructured text. After each simulation, they are converted to JSON to facilitate information extraction. The Analysis agent is further supported by the Python and Plotting agents, enabling flexible analysis workflows and generation of publication-ready figures.

Finally, the Force Field Fitting agent is responsible for diagnosing discrepancies between simulations and reference data and for developing or refining force-field parameters. The fitting harness is entered through a dedicated agent tool call and implemented as a state-machine controller. Each optimization attempt is structured into four phases: parameter selection, simulation, analysis, and a continue-or-terminate decision. The controller injects phase-specific instructions, delegates setup and execution to specialist agents, preserves every iteration separately, and limits the number of attempts. When the loop ends, it compacts the completed trajectory into a concise summary for subsequent reasoning.

In parallel, the force-field extraction pipeline enables the system to retrieve and process information from scientific publications. A Literature agent coordinates this process, delegating tasks to a Search agent that identifies relevant papers (using Semantic Scholar \cite{SI-ammar2018construction}) and a Reader agent that parses and interprets their content. Retrieved documents are automatically converted into structured representations using PyMuPDF4LLM \cite{SI-artifex2024pymupdf4llm}, enabling targeted extraction of key information such as force field parameters, simulation settings, and experimental data. A dedicated FF Writer agent translates extracted parameters into simulation-ready force field definitions, while an FF Adapter agent modifies or combines force fields to match specific use cases (e.g., adapting atom labels or combining parameter sets). While primarily used for force field construction, this pipeline can also extract other relevant information from the literature to support downstream simulation and analysis tasks.

\subsubsection{Agent Coordination and Reasoning}

RaspAssistant employs two distinct reasoning paradigms, depending on the complexity of the agent's role. High-level agents, such as the Supervisor, Setup, Analysis, and Literature agents, follow an iterative plan--execute--verify--replan strategy. These agents first construct a plan based on the task and available resources, execute subtasks sequentially, and explicitly verify the outcome of each step. Verification is performed using model-based self-evaluation, where the agent assesses whether the produced output satisfies the intended objective. If necessary, the plan is updated to account for errors or newly available information.

In contrast, lower-level agents follow the ReAct paradigm \cite{SI-yao2023react}, in which they iteratively select actions based on the current state and available tools. This approach is well-suited for more structured tasks such as file manipulation, simulation execution, or example retrieval.

The agents use largely general-purpose reasoning strategies and tool interfaces. Outside the phase-controlled fitting protocol (Section~\ref{SI-sec:fitting_harness}), agents combine high-level planning with flexible tool use rather than following a fixed end-to-end study workflow. Because they can create and revise complete input files rather than select only from schema-defined parameters, the architecture can accommodate input variants and task sequences that were not explicitly anticipated when the workflow was designed. It can also extend to previously unseen simulation types when the required executable, documentation, and examples are available. This design naturally extends to how failures are handled: instead of relying on predefined error-handling routines, agents diagnose issues based on the current state and iteratively adjust their strategy, for example by modifying inputs, selecting alternative tools, or revising intermediate steps. These capabilities support adaptation in the presence of incomplete information or unexpected outcomes.

\subsubsection{Tools, Memory, and Execution Environment}

Agents interact with their environment through a set of tools bound to the underlying language models. These tools provide capabilities such as file system interaction (e.g., reading, writing, and listing files), execution of simulations, viewing images, and code execution. This abstraction enables agents to operate on real simulation workflows while maintaining a consistent interface.

The system is executed on a local machine, where simulation software is accessed through a wrapped interface that mimics command-line execution. Molecular simulations were executed with RASPA 2.0.47. To reduce computational cost during evaluation, previously computed results can be reused when simulation inputs are identical, allowing the benchmark to avoid repeating equivalent calculations without changing the inputs or scientific evidence available to the agents. For batches of nominally identical long simulations, execution was limited to one representative per batch to reduce redundant computation. One run resumed after correction of an external RUPTURA configuration error, without additional scientific guidance. During workflow execution, human actions were limited to benchmark acceleration, resource management, and infrastructure maintenance, excluding scientific decisions or interpretation.

Elapsed time is reported as campaign wall-clock duration. Token use is the sum of input and output tokens. The table reports approximate token totals and approximate API-equivalent costs under the model prices used during execution.

RaspAssistant incorporates both persistent and transient memory mechanisms. Each agent can access a local memory stored as structured markdown files, which captures relevant past experiences such as simulation setups, parameter choices, and observed issues. Persistent memory was disabled between benchmark invocations, so each run began without benchmark-specific information from earlier runs. A shared per-run memory was initialized separately for each invocation to track the current workflow and facilitate coordination between agents.

A detailed overview of the agents, their roles, and capabilities is provided in Table~\ref{SI-tab:agents}. The table lists configured model families. Values in parentheses give case-study configurations where they differ from workflow reproduction. Literature-agent assignments are taken from the frozen reviewer configurations. Exact model identifiers recorded in saved runs are supplied in the supporting data package. A separate GPT-5 mini routing agent passes requests to the scientific Supervisor. PEV denotes plan--execute--verify--replan, and R/W/P/V denote file reading, file writing, Python execution, and vision.

The supporting data package includes benchmark prompts, agent and model configurations, software versions, stopping conditions, and simulation-reuse rules. Its artifact manifest links workflow records to the corresponding simulation inputs, outputs, and derived analysis files.

\begin{table*}[htbp]
\centering
\caption{Agent architectures, configured model families, and capabilities. Parentheses indicate case-study configurations where they differ from workflow reproduction.}
\label{SI-tab:agents}
\begin{tabularx}{\textwidth}{l l l c c c c X}
\toprule
\textbf{Agent} & \textbf{Type} & \textbf{Model} & \textbf{R} & \textbf{W} & \textbf{P} & \textbf{V} & \textbf{Other Tools} \\
\midrule

Supervisor   & PEV   & GPT-5.2   & \xmark & \xmark & \xmark & \xmark & Agent delegation, fitting-loop entry \\
Setup        & PEV   & GPT-5.2   & \checkmark & \checkmark & \xmark & \xmark & RASPA manual \\
Example      & ReAct & GPT-5 mini & \checkmark & \xmark & \xmark & \xmark & RASPA examples \\
File         & ReAct & GPT-5 mini & \checkmark & \checkmark & \checkmark & \xmark &  \\
Analysis     & PEV   & GPT-5.4   & \checkmark & \checkmark & \xmark & \checkmark & Output file parsing \\
Python       & ReAct & GPT-5.2   & \checkmark & \checkmark & \checkmark & \xmark & Output file parsing \\
Plotting     & ReAct & GPT-5.2 (5.1)   & \checkmark & \checkmark & \checkmark & \checkmark &  \\
RASPA        & ReAct & GPT-5.1   & \checkmark & \xmark & \xmark & \xmark & RASPA \\
RUPTURA      & ReAct & GPT-5.1 & \checkmark & \checkmark & \xmark & \checkmark & RUPTURA  \\
FF Fitter    & PEV   & GPT-5.2   & \checkmark & \checkmark & \checkmark & \checkmark & Output file parsing, FF fitting tool \\
\midrule

Literature   & PEV   & GPT-5 mini   & \checkmark & \xmark & \xmark & \xmark &  \\
Search       & ReAct & GPT-5 mini   & \checkmark & \xmark & \xmark & \xmark & Semantic Scholar\\
Reader       & ReAct & GPT-5.2   & \checkmark & \checkmark & \xmark & \checkmark & Paper reading \\
FF Writer    & ReAct & GPT-5.2   & \checkmark & \checkmark & \checkmark & \xmark & RASPA manual \\
FF Adapter   & ReAct & GPT-5.2   & \checkmark & \checkmark & \xmark & \xmark & RASPA manual \\

\bottomrule
\end{tabularx}
\end{table*}

\subsection{Workflow Reproduction Evaluation}
We evaluate RaspAssistant on workflow reproduction tasks using four representative case studies drawn from adsorption and separation studies in porous materials. These benchmarks are designed to assess the ability of the system to reproduce complete computational workflows from published studies, including simulation setup, execution, and analysis. In all cases, literature-derived force field parameters are provided as input, ensuring that the evaluation focuses on workflow automation and scientific reasoning rather than force field construction or fitting. Each case corresponds to a full end-to-end simulation pipeline derived from the literature, enabling controlled evaluation of agentic performance in realistic scientific scenarios.

Each benchmark provides (i) a natural-language task description, (ii) structural information (CIF files), (iii) experimental adsorption data where available, and (iv) pre-defined force field parameters extracted from the literature. The objective is to reproduce the computational workflow of the corresponding study as faithfully as possible, while allowing for minor variations in implementation.

As an architectural baseline, each benchmark was also executed once by a single GPT-5.4 agent. It received the same scientific request, input data, file workspace, simulation backends, tracking mechanisms, and the deduplicated union of the primitive tools available to the specialized simulation agents, but no delegation or handoff tools. The single agent maintained a dependency-ordered task list in one history. At an approximate occupancy of 70\% of its nominal 200,000-token context, older messages were replaced by a structured checkpoint while the original request and 12 most recent messages were retained, and the removed messages were archived. Baseline outputs were scored with the same rubric and evidence hierarchy as the multi-agent invocations. Because the multi-agent system uses a heterogeneous model mix, this is a descriptive architectural baseline rather than a model-, token-, or cost-matched ablation.

\subsubsection{Benchmark Cases}

\textbf{Study 1: CO$_2$/N$_2$ Separation in TAMOF-1 \cite{SI-gooijer2025tamof}.}
This case considers adsorption and separation of CO$_2$ and N$_2$ in the TAMOF-1 metal--organic framework. The workflow includes single-component adsorption isotherms, validation against experimental data, ideal adsorbed solution theory (IAST) calculations, mixture adsorption simulations, and breakthrough curve analysis. This benchmark evaluates the ability of the system to reproduce a complete adsorption-to-separation pipeline, including both equilibrium and dynamic simulations.

\textbf{Study 2: C$_2$H$_2$/CO$_2$ and C$_2$H$_2$/CH$_4$ Separation in Open-Metal-Site MOFs \cite{SI-luna2019acetylene}.}
This case focuses on adsorption and separation of acetylene, methane, and carbon dioxide in metal--organic frameworks with open metal sites. The workflow includes pure-component adsorption isotherms, validation against experimental data, thermodynamic and energetic analysis, mixture adsorption simulations, and selectivity evaluation. In addition, adsorption site analysis is performed to interpret molecular-scale binding mechanisms governing separation performance.

\textbf{Study 3: Metal Center Effects in M-MOF-74 \cite{SI-wasik2024impact}.}
This case investigates CO$_2$/H$_2$ separation across the M-MOF-74 series with different metal centers. The workflow includes adsorption isotherm calculations and binding geometry analysis, followed by validation against experimental isotherms and adsorption geometries. Ideal adsorbed solution theory (IAST) and mixture grand canonical Monte Carlo simulations are then used to evaluate separation performance, followed by breakthrough simulations. The final objective is to rank metal centers according to their separation efficiency.

\textbf{Study 4: Olefin/Paraffin Separation in Cation-Exchanged Zeolites \cite{SI-luna2020pi}.}
This case investigates olefin/paraffin separation in LTA- and FAU-type zeolites containing Na and/or Ca charge-balancing cations. The adsorbates considered are C$_3$--C$_6$ hydrocarbons, allowing the effect of chain length on separation performance to be assessed. The workflow includes calculation of Rosenbluth weights for flexible hydrocarbons, validation against pure-component propane/propene adsorption isotherms and hexane/hexene isobars, and mixture adsorption simulations to determine separation selectivities. In addition, adsorption site analysis is performed to examine how cation identity, zeolite topology, and hydrocarbon chain length influence molecular binding and separation behaviour. The final objective is to identify structure--composition trends that govern olefin/paraffin separation in cation-exchanged zeolites.

\subsubsection{Evaluation Metrics}

We score each workflow-reproduction run for workflow correctness, simulation validity, and analysis consistency. Table~\ref{SI-tab:workflow_metrics} summarizes the evaluation dimensions and weights. Detailed metric definitions and benchmark-specific workflow components are provided in Sections~\ref{SI-sec:workflow_evaluation} and \ref{SI-sec:benchmark_criteria}, respectively.

\begin{table}[htbp]
\centering
\caption{Workflow-reproduction evaluation dimensions and weights.}
\label{SI-tab:workflow_metrics}
\begin{tabularx}{\linewidth}{l X c}
\toprule
\textbf{Dimension} & \textbf{Definition} & \textbf{Weight} \\
\midrule
Workflow correctness & Completion of the expected workflow steps in the appropriate order and using appropriate methodological choices. & 40\% \\
Simulation validity & Physical and computational validity of the simulations, including parameter selection and convergence. & 40\% \\
Analysis consistency & Internal support and calibration of the scientific interpretation, including uncertainty and model limitations. & 20\% \\
\bottomrule
\end{tabularx}
\end{table}

\subsubsection{Workflow Scoring Procedure}

Each workflow-reproduction run is evaluated for workflow correctness, simulation validity, and analysis consistency. Each dimension is reported on a 0--10 scale. The overall score uses the 40/40/20 weighting in Equation~\ref{SI-eq:overall_score}.

\paragraph{Workflow correctness.}
Each benchmark is decomposed into weighted steps derived from the task specification and corresponding literature workflow. Steps receive full, partial, or zero credit according to their coverage and methodological appropriateness. Step weights $w_j$ and awarded fractions $c_j$ enter Equation~\ref{SI-eq:workflow_score}. The benchmark-specific steps and scoring anchors are provided in Sections~\ref{SI-sec:benchmark_criteria} and \ref{SI-sec:general_scoring_rules}, respectively.

\paragraph{Simulation validity.}
This metric assesses whether the simulations are physically meaningful and computationally sound. It considers the simulated systems and state points, force fields and electrostatics, periodic-cell and cutoff choices, sampling, units, output integrity, and the validity of downstream methods. Model--experiment disagreement is not itself penalized. A systematic issue is scored once according to its scope, and only dependent result families are affected.

\paragraph{Analysis consistency.}
This metric assesses whether the conclusions are internally correct, supported by the run's usable evidence, calibrated to model sensitivity and uncertainty, and based on an appropriate hierarchy of evidence. Agreement with the reference study and exact material rankings are not required. A cautious conclusion that the available results are inconclusive can therefore receive full credit, whereas unsupported precision or propagation of an invalid upstream model is penalized.

Sections~\ref{SI-sec:general_scoring_rules} and \ref{SI-sec:benchmark_criteria} specify the scoring anchors, result-family validity labels, non-double-penalization rules, and benchmark-specific workflow criteria.

The retained simulation inputs, completed outputs, derived analyses, and reports were scored by one human assessor using this rubric. Scores did not depend on numerical agreement with the reference publication.

\subsection{Force-Field Extraction Evaluation}

In addition to workflow reproduction, we evaluate the ability of RaspAssistant to extract and construct simulation-ready force fields from scientific literature. For this benchmark, we select eight representative force-field extraction cases spanning different classes of porous materials and interaction models. The goal is to assess whether the system can correctly identify, extract, and format force field parameters into RASPA-compatible input files.

We score force-field extraction for parameter correctness and RASPA readiness. Table~\ref{SI-tab:ff_metrics} summarizes the evaluation dimensions and weights. Section~\ref{SI-sec:ff_extraction_evaluation} defines the metrics and Table~\ref{SI-tab:ff_scope} lists the benchmark cases.

\begin{table}[htbp]
\centering
\caption{Force-field extraction evaluation dimensions and weights.}
\label{SI-tab:ff_metrics}
\begin{tabularx}{\linewidth}{l X c}
\toprule
\textbf{Dimension} & \textbf{Definition} & \textbf{Weight} \\
\midrule
Parameter correctness & Correctness and completeness of the extracted parameters relative to the reference parameter set. & 60\% \\
RASPA readiness & Fraction of generated force-field files that execute in RASPA without correction. & 40\% \\
\bottomrule
\end{tabularx}
\end{table}

Readiness was assessed by using the generated files in RASPA without manual correction. Parser errors, missing definitions, and inconsistent declared entry counts were treated as non-executable.

\subsection{Case Studies}

In addition to the structured benchmark tasks, we evaluate the system on two less prescriptive case studies. Whereas the benchmark tasks are designed to test whether RaspAssistant can reproduce predefined workflows with relatively explicit instructions, these case studies assess its ability to make useful decisions in a more open-ended setting. We therefore focus not only on the final numerical results, but also on the choices made by the system, including force field selection, validation strategy, simulation design, and interpretation of the resulting data. The two case studies are closely related to the workflow-reproduction benchmarks, but require the system to decide how to adapt the available information to a less fully specified research objective.

\subsubsection{Case Study 1: C$_2$H$_2$/CO$_2$ and C$_2$H$_2$/CH$_4$ Separation in Open-Metal-Site MOFs}

This case study examines whether RaspAssistant can identify and address limitations in a generic force field when applied to adsorption in metal--organic frameworks with open metal sites~\cite{SI-luna2019acetylene}. Instead of providing the final fitted force field from the reference study, we provide the initial force field used as the starting point in that work, consisting of UFF and Dreiding parameters for the framework and adsorbate definitions taken from prior literature. The system must therefore recognize that standard mixing rules are insufficient to describe the strong interaction between acetylene and open metal centres, diagnose the origin of the discrepancy with reference data, and propose an appropriate fitting strategy.

To keep the computational cost manageable, the study is restricted to two representative systems: IRMOF-1, which does not contain open metal sites, and Fe-MOF-74, which contains coordinatively unsaturated metal centres. This allows the system to distinguish between cases where the generic force field is expected to be adequate and cases where additional framework--adsorbate interactions are required. The evaluation focuses on whether RaspAssistant can identify the need for force field refinement, perform a controlled fitting procedure, and justify the resulting parameters in terms of adsorption behaviour and molecular interactions.

\subsubsection{Case Study 2: High-Throughput Screening of Pure-Silica Zeolites for CO$_2$ Adsorption}
\label{SI-sec:mfi_validation_metrics}

This case study evaluates whether RaspAssistant can perform an end-to-end screening workflow for CO$_2$ adsorption in zeolites. The supplied dataset comprised 198 structures from the IZA Database of Zeolite Structures~\cite{SI-baerlocher2007atlas,SI-iza2026database}. Of these, 197 passed automated input checks and were retained for screening. Experimental CO$_2$ adsorption isotherms in pure-silica MFI at several temperatures were also supplied~\cite{SI-garcia2009transferable}. Per-temperature mean absolute errors use linear-pressure interpolation of simulations at experimental points within the shared pressure domain. The aggregate MAE weights each temperature by its number of overlapping experimental points. Open simulation markers identify interquartile-range outliers in the archived block-drift diagnostic (\texttt{block\_drift\_iqr\_outlier}). These flagged points remain in the interpolation and error calculations. The task is to select a suitable force field for CO$_2$ adsorption in pure-silica zeolites, validate it against the available MFI experimental data, and then apply the selected force field to the accepted zeolite structures.

Unlike the workflow-reproduction benchmarks, the system is not given a fully specified simulation protocol. It must decide which candidate force fields to consider, how to compare simulated and experimental isotherms, and which criteria should be used to select the most reliable model. After validation, the chosen force field should be used to compute adsorption properties across the zeolite dataset under consistent simulation conditions. The resulting predictions are then analyzed to identify promising zeolite topologies and to relate adsorption performance to structural features such as pore size, topology, and adsorption capacity. This case study therefore tests the ability of RaspAssistant to combine literature-based force field selection, experimental validation, large-scale simulation setup, and post-processing into a complete materials screening workflow.

\section{Additional Interpretation and Research Context}

The force-field extraction and case study results further demonstrate the usefulness of this approach. The literature agents extracted simulation-ready force field parameters with near-perfect agreement to the reference data, showing that agentic systems can help address a common practical bottleneck in classical molecular simulation. In the open-metal-site MOF case study, RaspAssistant identified the failure of a generic force field, carried out a targeted fitting procedure, and used the resulting model for downstream mixture adsorption and site analysis. In the zeolite screening case study, the system selected a literature-based CO$_2$ force field through comparison with experimental MFI data and applied it consistently to a larger set of pure silica zeolites. Together, these examples show that the framework can support both reproduction of published workflows and less prescriptive simulation studies that require validation-based decision-making.

Together, these results position RaspAssistant as a system for autonomous execution of complete adsorption-simulation workflows after task specification and input provision. Its contribution lies in carrying validation, adaptation, and documentation across coupled simulation stages while retaining expert-defined criteria for physical adequacy.

Expanded research context is provided in Section~\ref{SI-sec:additional_context}. Here, flexibility refers to the range of scientific decisions the agents can make and revise across an adsorption study, beyond generating input files. Among the systems discussed in Section~\ref{SI-sec:additional_context}, we found no reported evaluation spanning force-field selection or adaptation, experimental validation, pure-component and mixture simulations, and process-level analysis through reproduction of complete published porous-materials adsorption and separation studies.

The final fitted parameters are compared with the literature force field in Table~\ref{tab:ff_comparison}. Although the final adsorption isotherm is very similar to the literature result, the fitted parameters differ substantially. In the literature model, the correction is distributed across several framework atom types, including framework oxygen, carbon, hydrogen, and the metal site. In contrast, the force field obtained by RaspAssistant concentrates the correction primarily in a much stronger Fe--C$_{\mathrm{C_2H_2}}$ interaction, with additional weaker oxygen contributions. This indicates that the system was able to find an effective parameterization for the target adsorption behaviour, while following a different parameterization strategy from the literature force field.

After fitting the C$_2$H$_2$ interactions, RaspAssistant proceeded with the remaining adsorption and separation analysis. The fitted force field was used to run mixture adsorption simulations and calculate C$_2$H$_2$/CO$_2$ and C$_2$H$_2$/CH$_4$ selectivities. As shown in Figure~\ref{fig:case1_selectivity}, the resulting selectivities are in overall good agreement with the values reported in the reference study, reproducing both the magnitude and the pressure dependence for Fe-MOF-74 and IRMOF-1. Some deviations are observed, particularly in pressure regions where the uptake of the less strongly adsorbed component, CH$_4$ or CO$_2$, is very small. In these cases, the selectivity is highly sensitive to small absolute differences in loading, so modest discrepancies in mixture uptake can lead to larger relative differences in the selectivity. Overall, this comparison shows that the fitted force field is not only able to reproduce the pure-component adsorption behaviour used during fitting, but also remains suitable for the downstream mixture separation analysis.

Finally, RaspAssistant performed an adsorption site analysis to interpret the molecular origin of the observed separation behaviour. The reference study visualized adsorption sites using spatial histograms, whereas RaspAssistant quantified the same effect through nearest Fe distance distributions. These distributions convey the same physical picture: C$_2$H$_2$ is strongly localized near the open metal sites, while CO$_2$ and CH$_4$ are predominantly located farther away in the pore space. As shown in Figure~\ref{fig:case1_rdf}, C$_2$H$_2$ exhibits a sharp peak at short Fe distances in both mixtures, indicating preferential binding near the Fe sites. In contrast, CO$_2$ and CH$_4$ show broader distributions at larger distances, consistent with weaker localization near the open metal sites. Together with the isotherm and selectivity results, this analysis shows that RaspAssistant was able to proceed beyond force field fitting and mixture simulations to provide a molecular interpretation of the predicted separation behaviour.

RaspAssistant also examined relationships between the screening results and simple structural descriptors. Figure~\ref{fig:case2_density_outliers} shows the relationship between framework density and simulated CO$_2$ uptake at 1 kPa. Framework density showed a weak positive association with uptake at 1 kPa (Spearman $\rho=0.284$), while GIS, JBW, SIV, ATN, and YUG were positive residual outliers relative to this trend. Because the retained descriptor set did not include pore size, accessible volume, or channel-connectivity measures, these residuals cannot be assigned to a specific pore-scale mechanism.

The descriptor analysis also revealed a limitation of the workflow. RaspAssistant could use simple crystallographic and density-based quantities, but did not have direct access to more informative geometric pore descriptors such as accessible pore volume, pore-limiting diameter, largest included sphere, or helium void fraction. As a result, the system could identify density-normalized outliers, but could not fully explain them in terms of pore size or accessible volume. A natural extension would be to connect the workflow to a geometric analysis package such as Zeo++~\cite{SI-willems2012algorithms}. This would give the agents direct access to pore descriptors during post-processing, allowing future screening workflows to relate adsorption performance more directly to channel geometry, pore accessibility, and confinement effects.

The agentic systems discussed above cover computational chemistry, materials simulation, and screening~\cite{SI-zou2025agente,SI-gustin2025agente,SI-choi2026agente,SI-perez2026agente,SI-wang2025dreams,SI-liu2025vaspilot,SI-rothfarb2026master,SI-chaudhari2026matsciagent,SI-yang2026quasar,SI-lee2026simmof,SI-ghafarollahi2025atomagents,SI-shi2025mdagent,SI-wang2026agentmd}. RaspAssistant is evaluated on the continuity of an adsorption study from model selection through process-level interpretation. Direct numerical comparison is not appropriate because these systems address different tasks and use different evaluation criteria. The present results instead demonstrate that a hierarchical agent architecture can preserve scientific context across several coupled tools and stages of a porous-materials workflow.

The benchmark also highlights experiment-cost awareness as an opportunity for further development. In Acetylene Run~1, the system adopted a conservative strategy of requesting 20--30 long replicas for individual conditions without access to a prospective estimate of queue size or computational cost. We limited execution to one copy per launch because equilibrated copies were expected to provide little additional information. This example suggests a practical route to more efficient automation: future versions could provide planning agents with predicted queue size and simulation cost, configurable budgets, and explicit uncertainty objectives before launching replicas.

One direction for future development is to broaden the set of domain-specific tools available to the agents. Integrating geometric characterization packages such as Zeo++~\cite{SI-willems2012algorithms} would provide direct access to descriptors including accessible pore volume, pore-limiting diameter, and largest included sphere, while charge-assignment and charge-equilibration methods could extend automated structure preparation. Interfaces to electronic-structure codes, force-field parameterization and validation utilities, and systematic convergence and uncertainty-analysis tools could further support model construction and verification. More broadly, the modular architecture provides a route for incorporating these and other capabilities as required by new materials classes and simulation workflows. These additions would strengthen reproducibility and resource awareness while preserving the role of expert judgement in defining acceptable physical approximations.

\subsection{Additional Workflow and Force-Field Extraction Details}

Adsorption and separation studies typically couple several simulation stages. A typical workflow may include pure-component adsorption isotherms, validation against experimental data, mixture-performance evaluation using Ideal Adsorbed Solution Theory (IAST) \cite{SI-myers1965iast,SI-walton2015iast50,SI-krishna2021iastreliability} or direct mixture simulations, and process-level analysis through breakthrough simulations \cite{SI-sharma2023ruptura,SI-hassan2026aim}. Tools such as pyIAST \cite{SI-simon2016pyiast}, RUPTURA \cite{SI-sharma2023ruptura}, and AIM \cite{SI-hassan2026aim} have lowered the barrier for individual stages of this workflow, but they do not by themselves automate the complete process from literature-grounded model construction to validated scientific interpretation.

Executable rates are slightly lower despite the high IoU scores. These failures are not caused by incorrect parameter values or syntax errors, but by inconsistencies in global bookkeeping. RASPA force field files require the total number of interactions or atoms to be declared at the top of the file, consistent with the number of entries defined below. The observed failures arise when this count is incorrect, even though the interaction definitions themselves are accurate. These files cannot be executed until the declared entry counts are corrected.

The cost of force-field extraction remains low across all benchmarks. Higher costs occur when parameters are distributed across multiple sources (C$_x$H$_x$, CO$_2$, N$_2$, O$_2$ in zeolites), requiring additional retrieval steps, or when the force field contains a large number of parameters, as in the combined MOF case.
\subsection{Open-Ended Case Studies}
Unlike the workflow-reproduction benchmarks, these studies required higher-level modelling decisions rather than reproduction of a prescribed target workflow. Case Study~1 required the system to diagnose and refine an inadequate acetylene interaction model, whereas Case Study~2 required selection and experimental validation of a CO$_2$ force field before large-scale zeolite screening.

\subsection{Supporting Data Package}
\label{SI-sec:data_package}

The data package supporting this study is available on Zenodo and contains chronological workflow records in machine-readable JSON and self-contained HTML. Each record presents the original task, the top-level scientific interaction and its tool calls and responses, the instructions supplied for scientific delegations, and the outcome of each delegated task. For every workflow-reproduction invocation, the package also includes the final report, the scientific files cited in that report, and the corresponding evaluation record. The package additionally provides the source data underlying the reported figures and tables, the final force-field definitions and numerical result summaries for the two open-ended case studies, and representative simulation inputs with source-access instructions for external structure files. An artifact manifest links these records to the corresponding scientific files.

\section{Evaluation Metrics}

Workflow reproduction and force-field extraction were assessed using the following rubric. Workflow-reproduction scores are reported on a 0--10 scale, whereas force-field extraction metrics are reported on a 0--1 scale unless stated otherwise.

\subsection{Workflow-Reproduction Evaluation}
\label{SI-sec:workflow_evaluation}

Each workflow-reproduction run is evaluated for workflow correctness, simulation validity, and analysis consistency. The three dimensions contribute 40, 40, and 20 raw points, respectively. Dimension scores are reported on a 0--10 scale, and the weighted overall score is
\begin{equation}
S_{\mathrm{overall}} = 0.40S_{\mathrm{workflow}} + 0.40S_{\mathrm{validity}} + 0.20S_{\mathrm{analysis}}.
\label{SI-eq:overall_score}
\end{equation}

\subsubsection{General Scoring Rules}
\label{SI-sec:general_scoring_rules}

The assessment is based on simulation inputs, completed raw outputs, derived data, plots, and the final report. Completed raw simulation output and effective settings reported by the engine take precedence. Next are simulation inputs, structures, force-field files, and source-data metadata, followed by derived tables, scripts, plots, and analysis records. The final scientific report has the lowest priority. Effective simulation behaviour is assessed rather than the literal presence of a keyword in an input file.

Numerical agreement with experiment or with the reference publication is not itself rewarded or penalized. The supplied force fields and process models can be sensitive or systematically inaccurate, including for breakthrough calculations. The evaluation instead asks whether the comparison was performed on a consistent basis and whether disagreement was identified and interpreted with appropriate caution. Exact material rankings are likewise not required when the differences are within uncertainty or depend on the mixture model.

\paragraph{Workflow correctness.}
Each benchmark is decomposed into weighted atomic steps. A step normally receives full credit ($c_j=1$), partial credit ($c_j=0.5$), or zero credit ($c_j=0$). Values of 0.25 or 0.75 are permitted only for a documented borderline case. The score is
\begin{equation}
S_{\mathrm{workflow}} = 10\frac{\sum_j w_j c_j}{\sum_j w_j},
\label{SI-eq:workflow_score}
\end{equation}
where $w_j$ is the weight of workflow step $j$. Full credit requires the requested coverage and a suitable method. Partial credit represents a meaningful but incomplete or methodologically qualified attempt. Zero credit represents an absent step or one performed on the wrong central system or conditions.

\paragraph{Simulation validity.}
Simulation validity is divided into the seven criteria in Table~\ref{SI-tab:validity_criteria}. Each criterion is awarded a fraction $v_k$ using the same full/partial/zero anchors, and the 40-point validity contribution is $P_{\mathrm{validity}}=\sum_k u_kv_k$, where $u_k$ is the listed point value. Model--experiment disagreement is not a validity error. Examples of validity errors are inconsistent units, wrong state points, ignored charges, inappropriate simulation settings, incomplete jobs, and use of a failed upstream model without qualification.

\begin{table}[htbp]
\centering
\caption{Simulation-validity criteria used for every workflow-reproduction benchmark.}
\label{SI-tab:validity_criteria}
\small
\begin{tabularx}{\linewidth}{l X r}
\toprule
\textbf{ID} & \textbf{Full-credit requirement} & \textbf{Points} \\
\midrule
V1 & Correct structures, species, temperatures, pressures, and compositions for the requested result families. & 6 \\
V2 & Supplied force fields and charge convention are used effectively. Required specialized interactions are retained. Effective electrostatics are confirmed from output. & 7 \\
V3 & Appropriate supercell replication and cutoff settings are used for the benchmark. & 6 \\
V4 & Equilibration and production are credible, and convergence, uncertainty, or an equivalent stability check is demonstrated. & 6 \\
V5 & Pressure, loading, heat, geometry, and normalization conventions are reconciled and comparisons are restricted to overlapping domains. & 7 \\
V6 & Inputs, completed outputs, warnings, derived values, and report claims are traceable. & 4 \\
V7 & Fits, IAST or direct-mixture calculations, site analyses, and breakthrough dependencies are checked before quantitative use. & 4 \\
\bottomrule
\end{tabularx}
\end{table}

Nominal replicas count as independent samples only when their seeds or outputs demonstrate independence. For V5, using an ideal fugacity coefficient at 2.5--4~MPa is not a hard failure because the benchmark tasks do not prescribe an equation of state. A small deduction is appropriate only when the assumption is silent and exact high-pressure magnitudes or rankings are emphasized.

\paragraph{Analysis consistency.}
The 20-point analysis contribution is the sum of the four criteria in Table~\ref{SI-tab:analysis_criteria}. Interpretation is graded against the run's own scientifically usable evidence. A well-supported conclusion that the available results are inconclusive or model-sensitive can receive full credit.

\begin{table}[htbp]
\centering
\caption{Analysis-consistency criteria used for every workflow-reproduction benchmark.}
\label{SI-tab:analysis_criteria}
\small
\begin{tabularx}{\linewidth}{l X r}
\toprule
\textbf{ID} & \textbf{Full-credit requirement} & \textbf{Points} \\
\midrule
I1 & Species, equations, signs, and reported values are internally correct. & 5 \\
I2 & Claims follow the run's usable outputs. A conclusion of ``inconclusive'' is accepted when supported. & 5 \\
I3 & Small denominators, force-field sensitivity, conflicting models, and sampling limitations are recognized. & 5 \\
I4 & Experimental comparison, pure fits, IAST, direct mixtures, breakthrough, and site evidence are weighted according to their reliability. Mechanistic claims have direct support. & 5 \\
\bottomrule
\end{tabularx}
\end{table}

\paragraph{Result-family status and non-double-penalization.}
Each major result family is labelled valid, qualified, or invalid. A valid family supports the claims made. A qualified family supports only restricted or qualitative claims with a stated limitation. An invalid family is not usable for its claimed purpose. A systematic issue is deducted once according to its scope. Dependent analyses inherit the status of the faulty input, but do not each incur a repeated numerical penalty. For example, a complete pure-isotherm grid can receive workflow credit despite an invalid unit conversion. The conversion is penalized under validity, and full analysis credit remains possible when the limitation is identified and no unsupported claim is made. Conversely, an IAST--GCMC disagreement is not itself a failure, but ignoring it when making precise breakthrough claims reduces analysis consistency.

Completion is established from retained inputs and completed raw outputs. Summary labels or status fields are not sufficient when requested result families are absent, while a documented repair can establish completion when it produces and verifies the missing outputs.

\paragraph{Single-agent baseline.}
Each workflow-reproduction benchmark was additionally run once with a single GPT-5.4 agent. The baseline received the same task, input data, workspace, simulation interfaces, tracking and memory mechanisms, and the deduplicated union of the primitive tools available to the specialized simulation agents, but it had no delegation or handoff tools. All actions and their results were therefore retained in a single persistent conversation history. At an approximate occupancy of 70\% of a nominal 200,000-token context window, and after completed tasks, older messages were replaced by a structured checkpoint while retaining the original request and 12 recent messages. Removed messages were archived verbatim. Failed attempts remain part of the completion record even when later autonomous repair completed the scientific grid. The same 40/40/20 rubric, evidence hierarchy, result-family status rules, and non-double-penalization procedure were used for the baseline and multi-agent runs. The comparison is not model-matched and is interpreted descriptively.

\subsection{Workflow-Reproduction Task Prompts}
\label{SI-sec:task_prompts}

The following natural-language prompts were supplied to the Supervisor Agent for the four workflow-reproduction benchmarks. Original spelling, notation, and task wording are retained.

\subsubsection{Study 2: C2H2/CO2 and C2H2/CH4 separation}

\begin{BenchmarkPrompt}
We want to investigate C2H2/CO2 and C2H2/CH4 separation using different MOF materials (Cu-BTC, Fe-MOF-74, IRMOF-1 and PCN-16).
We want to understand which material perfoms best at these separations, and how the performance changes with pressure.
---
Data:
- input_data/structures contains CIF files for Cu-BTC, Fe-MOF-74, IRMOF-1 and PCN-16.
- input_data/exp_data/isotherms contains experimental pure component isotherms for CO2, C2H2 and CH4 in these materials at various temperatures.
- input_data/exp_data/isosteric_heat contains experimental isosteric heat of adsorption data for CO2, C2H2 and CH4 at various temperatures.
- input_data/exp_data/binding_geometry contains experimental binding geometry data for CO2 and C2H2 in Fe-MOF-74 at 298K.
- input_data/forcefield contains all necesarry force field files per material, including any necessary pseudo atom files and mixing rules.

Use this data to validate the results from simulations to understand the quality of the results.
---
Task: Measure the separation performance of the various MOF materials for C2H2/CO2 and C2H2/CH4 separation. Specifically, we want to:

- Validate the simulation methods and results against the experimental data. This includes:
    - isotherms for pure CO2, C2H2 and CH4 at various temperatures for the different materials
    - isosteric heat of adsorption for CO2, C2H2 and CH4 at various temperatures for the different materials
    - binding geometry for CO2 and C2H2 at 298K

- Investigate and compare pure component adsorption behaviour for CO2, C2H2 and CH4 at 318K for the different materials.
  How do host-guest and guest-guest interactions influence the pure component adsorption behaviour of these molecules in the different materials, and how does this relate to the separation performance?

- Measure and compare selectivity (S) for an equimolar mixture of C2H2/CO2 and C2H2/CH4 at 298K using mixture isotherms. We want to know how this metric compares the different materials at various pressures.

    S = (x_CO2/ x_N2) / (y_CO2/y_N2), where x is the mole fraction in the adsorbed phase and y is the mole fraction in the gas phase.
    The pressure range should cover the same range as experimental data.

- For the best MOF, analyze the adsorption sites for the equimolar mixture of C2H2/CO2 and C2H2/CH4 at 298K and 10 bar, and understand how this relates to the separation performance.

Notes:
The pressures at which you evaluate should cover the same range as the experimental data (you do not need to use the same pressures).
Any simulation and results folders you create should be placed in the sims folder.
---
Output:
At the end you should create a report on the following:
- The selectivity of the different materials for C2H2/CO2 and C2H2/CH4 separation at 298K, and how this metric changes with pressure.
- The predicted adsorption sites for CO2 and C2H2 in an equimolar mixture for the best MOF, and how this relates to the separation performance.
- A comparison of the simulation results with the experimental data, including any insights on the quality of the simulations and potential reasons for discrepancies.
- Paths to any files you created with the results, including isotherm data, fitted isotherm parameters, IAST predictions, and adsorption site data and plots.
\end{BenchmarkPrompt}

\noindent\textit{Reading note:} Original request notation is retained. The mixture species and evaluation criteria are specified in Section~\ref{SI-sec:criteria_acetylene}.

\subsubsection{Study 3: M-MOF-74 CO2/H2 separation}

\begin{BenchmarkPrompt}
We want to investigate CO2/H2 separation for M-MOF-74 with various metal centres (Co, Cu, Fe, Mn, Ni, Zn).
Specifically, we want to know how the perfromance of different metal centres affects the separation performance for CO2/H2 separation.
---
Data:
- input_data/structures contains CIF files for M-MOF-74 for different metal centres (Co, Cu, Fe, Mn, Ni, Zn). Use the charges from the CIF files.
- input_data/exp_data/isotherms contains experimental pure component isotherms for CO2 and H2 in M-MOF-74, at 77K for H2 and 298K for CO2. Different units are used for CO2 and H2.
- input_data/exp_data/isosteric_heat contains experimental isosteric heat of adsorption data for CO2 and H2 in M-MOF-74, at 77K for H2 and 298K for CO2. Different units are used for CO2 and H2.
- input_data/exp_data/binding_geometry contains experimental binding geometry data for CO2 and H2 in M-MOF-74 at 298K.
- input_data/forcefield contains all necesarry force field files and adsorbate definitions for CO2 and H2.

Use this data to validate the results from simulations to understand the quality of the results.
---
Task: Measure the separation performance of M-MOF-74 for CO2/H2 separation and rank the different metal centres based on their performance. Specifically, we want to:

- Validate the simulation methods and results against the experimental data. This includes:
    - isotherms for pure CO2 and H2 at 298K for CO2 and 77K for H2
    - isosteric heat of adsorption for CO2 and H2 at 298K for CO2 and 77K for H2
    - binding geometry for CO2 and H2 at 298K

- Determine which metal centre performs best for CO2/H2 separation, at 298K, 2.5MPa, column void fraction of 0.4, initial velocity 0.006791 and a length of 0.065m.
  Show how different CO2 mixture ratios (0.1/0.9, 0.2/0.8, 0.3/0.7, 0.4/0.6, 0.5/0.5 and 0.9/0.1 CO2/H2) affect the separation performance.

- Investigate where adsorption occurs (adsorption sites) in an equimolar CO2/H2 mixture at 4000kPa for different metal centres, and how this relates to the separation performance.

Notes:
The pressures at which you evaluate should cover the same range as the experimental data (you do not need to use the same pressures).
Any simulation and results folders you create should be placed in the sims folder.
---
Output:
At the end you should create a markdown report on the following:
- The separation performance of M-MOF-74 for CO2/H2 separation at the specified conditions, and how this changes with different metal centres and mixture ratios.
- The predicted adsorption sites for CO2 and H2 in an equimolar mixture for different metal centres, and how this relates to the separation performance.
- A comparison of the simulation results with the experimental data, including any insights on the quality of the simulations and potential reasons for discrepancies.
- Paths to any files you created with the results, including isotherm data, fitted isotherm parameters, IAST predictions, and adsorption site data and plots.
\end{BenchmarkPrompt}

\subsubsection{Study 1: TAMOF-1 CO2/N2 separation}

\begin{BenchmarkPrompt}
We want to investigate CO2/N2 separation for TAMOF-1.
Specifically, we want to know at which conditions (pressure, temperature, mixture ratio) TAMOF-1 performs best for separating CO2 from N2.
---
Data:
- input_data contains the CIF file for TAMOF-1 (Use the charges from the CIF file). Helium void fraction: 0.427
- input_data/exp_data/isotherms contains experimental pure component isotherms for CO2 and N2 in TAMOF-1 at 298K, 323K, and 353K.
- input_data/exp_data/breakthrough/CO2 contains experimental breakthroughs for single component CO2 in He carrier gas at various conditions (see readme for details).
- input_data/exp_data/breakthrough/CO2_N2 contains experimental breakthroughs for a mixture of CO2 and N2 in He carrier gas at a single condition (see readme for details).
- input_data/forcefield contains all necesarry force field files and adsorbate definitions for CO2 and N2.

Use this data to validate the results from simulations to understand the quality of the results.
---
Task: Measure the separation performance of TAMOF-1 for CO2/N2 separation using molecular simulations, at 298K, 323K, and 353K. Specifically, we want to:

- Validate the simulation methods and results against the experimental data. This includes:
    - isotherms for pure CO2 and N2 at 298K, 323K, and 353K
    - single component breakthroughs for CO2 at various conditions
    - mixture breakthrough for CO2/N2 at a single condition

- measure selectivity (S) and trade off selectivity and uptake (TSN). We want to know how these metrics change with pressure, temperature, and mixture ratio.

    S = (x_CO2/ x_N2) / (y_CO2/y_N2), where x is the mole fraction in the adsorbed phase and y is the mole fraction in the gas phase.
    TSN = ln(S) * q_CO2, where q_CO2 is the uptake of CO2 in mol/kg, and S is the selectivity for theose conditions.
    For S and TSN, the mixture ratios should be 0.1/0.9, 0.3/0.7, and 0.5/0.5 and 0.7/0.3 (CO2/N2). The pressure range should cover the same range as experimental data.

- predict breakthrough curves for 10
    Conditions for breakthrough curve predictions: 1 bar, 298K, 0.005 m/s, column length 5.8 cm and column void fraction 0.58.

Notes:
The pressures at which you evaluate should cover the same range as the experimental data (you do not need to use the same pressures).
Any simulation and results folders you create should be placed in the sims folder.
---
Output:

At the end you should create a report on the following:
- The selectivity and TSN of TAMOF-1 for CO2/N2 separation at the specified conditions, and how these metrics change with pressure, temperature, and mixture ratio.
- The predicted breakthrough curves for 10
- A comparison of the simulation results with the experimental data, including any insights on the quality of the simulations and potential reasons for discrepancies.
- Paths to any files you created with the results, including isotherm data, fitted isotherm parameters, IAST predictions, and breakthrough curve data and plots.
\end{BenchmarkPrompt}

\noindent\textit{Reading note:} The original request contains an incomplete feed-list line. The four prediction feeds used in evaluation are specified in Section~\ref{SI-sec:criteria_tamof}.

\subsubsection{Study 4: Olefin/paraffin separation in zeolites}

\begin{BenchmarkPrompt}
We want to investigate olefin/paraffin separation using different zeolite materials.
We want to understand which material perfoms best at these separations, and how chain length affects the performance.
---
Data:
- input_data/structures contains CIF files for zeolites with LTA topolgy (CaA, LTA5A, ITQ-29) and FAU topology (FAU_Si, FAU-2, NaY, NaX, CaX). For zeolites with cations, restart files with their positions are also included.
- input_data/exp_data/isotherms contains experimental pure component isotherms for propane and propene for LTA5A and NaY.
- input_data/exp_data/isobars contains experimental pure component isobars for hexane and 1-hexene for LTA5A and NaY.
- input_data/forcefield contains all necesarry force field files per material, including any necessary pseudo atom files and mixing rules.

Use this data to validate the results from simulations to understand the quality of the results.
---
Task: Measure the separation performance of the various zeolites for olefin/paraffin separation for C3-C6 hydrocarbons.:

- Validate the simulation methods and results against the experimental data. This includes:
    - isotherms for pure propane and propene for LTA5A and NaY at 298K.
    - isobars for pure hexane and 1-hexene for LTA5A and NaY at different pressures.

- Investigate the influence of framework and type/amount of cations:
    - Calculate average occupancy maps (histograms of adsorption positions) for the pure component adsorption behaviour of propene at 298K, 100kPA for zeolites of the FAU topology.

    - For all frameworks, measure and compare selectivity (S) for an equimolar mixture of olefins and paraffins for C3-C6 hydrocarbons at 298K and 100kPa, and investigate the influence of the chain length.
      S = (x_olefin/ x_paraffin) / (y_olefin/y_paraffin), where x is the mole fraction in the adsorbed phase and y is the mole fraction in the gas phase.
      The pressure range should cover the same range as experimental data.

Notes:
The pressures at which you evaluate should cover the same range as the experimental data (you do not need to use the same pressures).
Any simulation and results folders you create should be placed in the sims folder.
---
Output:
At the end you should create a report on the following:
- The selectivity of the different materials for olefin/paraffin separation for C3-C6 hydrocarbons at 298K, and how this metric changes with pressure.
- The influence of type and amount of cations on the adsorption behaviour of propene at 298K, 100kPA for zeolites of the FAU topology, using average occupancy maps.
- The influence of chain length on the separation performance by comparing selectivity for C3-C6 hydrocarbons at 298K, 100kPA.
- A comparison of the simulation results with the experimental data, including any insights on the quality of the simulations and potential reasons for discrepancies.
- Paths to any files you created with the results, including isotherm data, fitted isotherm parameters, IAST predictions, and adsorption site data and plots.
\end{BenchmarkPrompt}

\subsection{Benchmark-Specific Scientific Criteria}
\label{SI-sec:benchmark_criteria}

\subsubsection{TAMOF-1 CO$_2$/N$_2$ separation}
\label{SI-sec:criteria_tamof}
Table~\ref{SI-tab:tamof_workflow} defines the 40-point TAMOF-1 workflow. The adsorption helium void fraction is 0.427, whereas the predicted breakthrough column void fraction is 0.58. Experimental He-carrier breakthrough cases must be distinguished from the requested binary CO$_2$/N$_2$ predictions. Direct mixture GCMC may be representative rather than exhaustive when IAST supplies the complete feed--temperature--pressure grid. Accurate breakthrough timing is not required. The score reflects whether the supplied experiments were compared correctly and whether disagreement was acknowledged.

\begin{table}[htbp]
\centering
\caption{Workflow criteria for TAMOF-1 CO$_2$/N$_2$ separation~\cite{SI-gooijer2025tamof}.}
\label{SI-tab:tamof_workflow}
\small
\begin{tabularx}{\linewidth}{l X r}
\toprule
\textbf{ID} & \textbf{Expected workflow step} & \textbf{Points} \\
\midrule
TAM-W1 & Inspect the structure, charges, experimental metadata, and distinct adsorption and column void fractions. & 3 \\
TAM-W2 & Run pure CO$_2$ and N$_2$ GCMC at 298, 323, and 353~K over the experimental pressure range. & 6 \\
TAM-W3 & Compare unit-matched pure simulations with experiment over the overlapping domains. & 5 \\
TAM-W4 & Calculate selectivity and TSN over all four feeds, three temperatures, and the pressure range. & 7 \\
TAM-W5 & Use IAST and representative direct mixture GCMC to assess mixture-model reliability. & 4 \\
TAM-W6 & Compare breakthrough with the supplied pure-CO$_2$/He and CO$_2$/N$_2$/He experiments. & 6 \\
TAM-W7 & Predict 10, 15, 20, and 25\% CO$_2$ in N$_2$ at 1~bar, 298~K, 0.005~m~s$^{-1}$, 5.8~cm, and column void fraction 0.58. & 5 \\
TAM-W8 & Synthesize the results, limitations, and evidence paths in an auditable report. & 4 \\
\bottomrule
\end{tabularx}
\end{table}
\FloatBarrier

The use of direct-mixture checks reflects known limitations of IAST in heterogeneous adsorbents~\cite{SI-krishna2021iastreliability}. Interpretation should distinguish selectivity from TSN, recognize instability when N$_2$ uptake is very small, and downgrade IAST or breakthrough conclusions when direct mixture GCMC or experimental comparison exposes limitations. No exact selectivity or breakthrough agreement is prescribed.

\begin{figure}[htbp]
\centering
\includegraphics[width=0.72\linewidth]{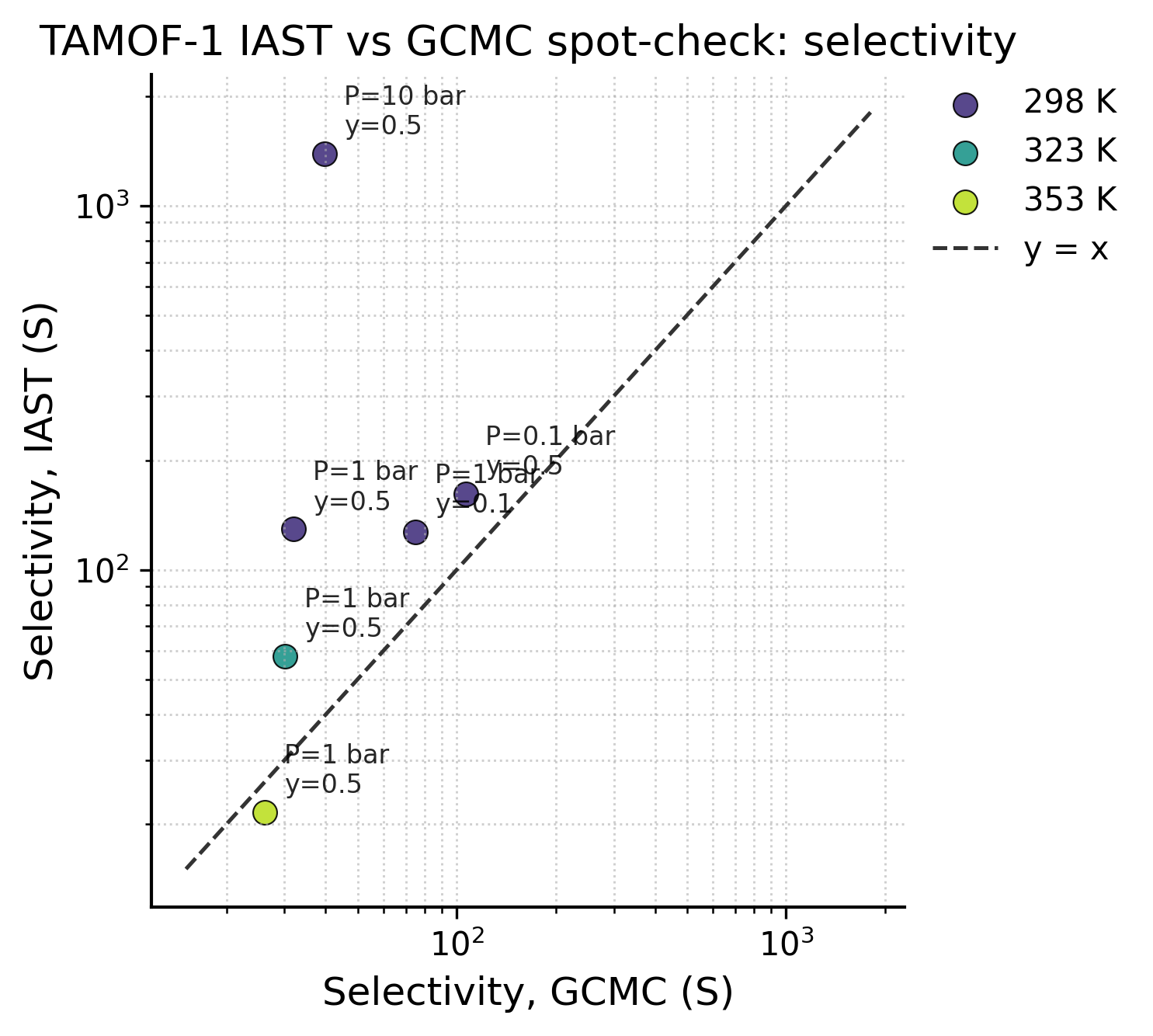}
\caption{TAMOF-1 diagnostic from Study~1, multi-agent Run~3, comparing IAST and direct-mixture GCMC selectivities for the benchmark based on Ref.~\cite{SI-gooijer2025tamof}. Deviations from the parity line motivate treating IAST-derived selectivities and downstream breakthrough predictions as qualified rather than requiring numerical agreement.}
\label{SI-fig:si_tamof_iast_gcmc}
\end{figure}
\FloatBarrier

\subsubsection{C$_2$H$_2$/CO$_2$ and C$_2$H$_2$/CH$_4$ separation}
\label{SI-sec:criteria_acetylene}
Table~\ref{SI-tab:acetylene_workflow} defines the 40-point open-metal-site-MOF workflow. The benchmark covers Cu-BTC, Fe-MOF-74, IRMOF-1, and PCN-16 with their supplied material-specific force fields. Native experimental temperatures are used for validation, 318~K for the pure-component mechanistic study, and 298~K for mixture and site analyses. The supplied task prompt uses generic CO$_2$/N$_2$ labels in its selectivity expression. For this benchmark, the expression applies to C$_2$H$_2$/CO$_2$ and C$_2$H$_2$/CH$_4$.

\begin{table}[htbp]
\centering
\caption{Workflow criteria for C$_2$H$_2$/CO$_2$ and C$_2$H$_2$/CH$_4$ separation~\cite{SI-luna2019acetylene}.}
\label{SI-tab:acetylene_workflow}
\small
\begin{tabularx}{\linewidth}{l X r}
\toprule
\textbf{ID} & \textbf{Expected workflow step} & \textbf{Points} \\
\midrule
ACE-W1 & Inventory the four structures, material-specific force fields, and experimental states. & 3 \\
ACE-W2 & Validate pure C$_2$H$_2$, CO$_2$, and CH$_4$ isotherms at their native experimental temperatures. & 6 \\
ACE-W3 & Validate adsorption heats using compatible loading bases and overlapping domains. & 4 \\
ACE-W4 & Validate Fe-MOF-74 CO$_2$ and C$_2$H$_2$ binding geometries at 298~K. & 3 \\
ACE-W5 & Perform the 318~K pure-component study with host--guest and guest--guest evidence. & 6 \\
ACE-W6 & Run equimolar mixture GCMC for both gas pairs at 298~K over the pressure range. & 8 \\
ACE-W7 & Select a candidate transparently and analyze both 298~K, 10~bar mixture trajectories or adsorption sites. & 6 \\
ACE-W8 & Synthesize the comparison, uncertainty, limitations, and evidence paths. & 4 \\
\bottomrule
\end{tabularx}
\end{table}
\FloatBarrier

Site analysis must use periodic distances, and open-metal-site or host--guest causal claims require trajectory, occupancy, geometry, or energy evidence. Replicas are required only if replica-based uncertainty is claimed. Identical replicas count as one effective sample. No particular winning material is prescribed, and large low-pressure selectivities should be qualified when the weak-component loading is statistically indistinguishable from zero.

\begin{figure}[htbp]
\centering
\includegraphics[width=0.98\linewidth]{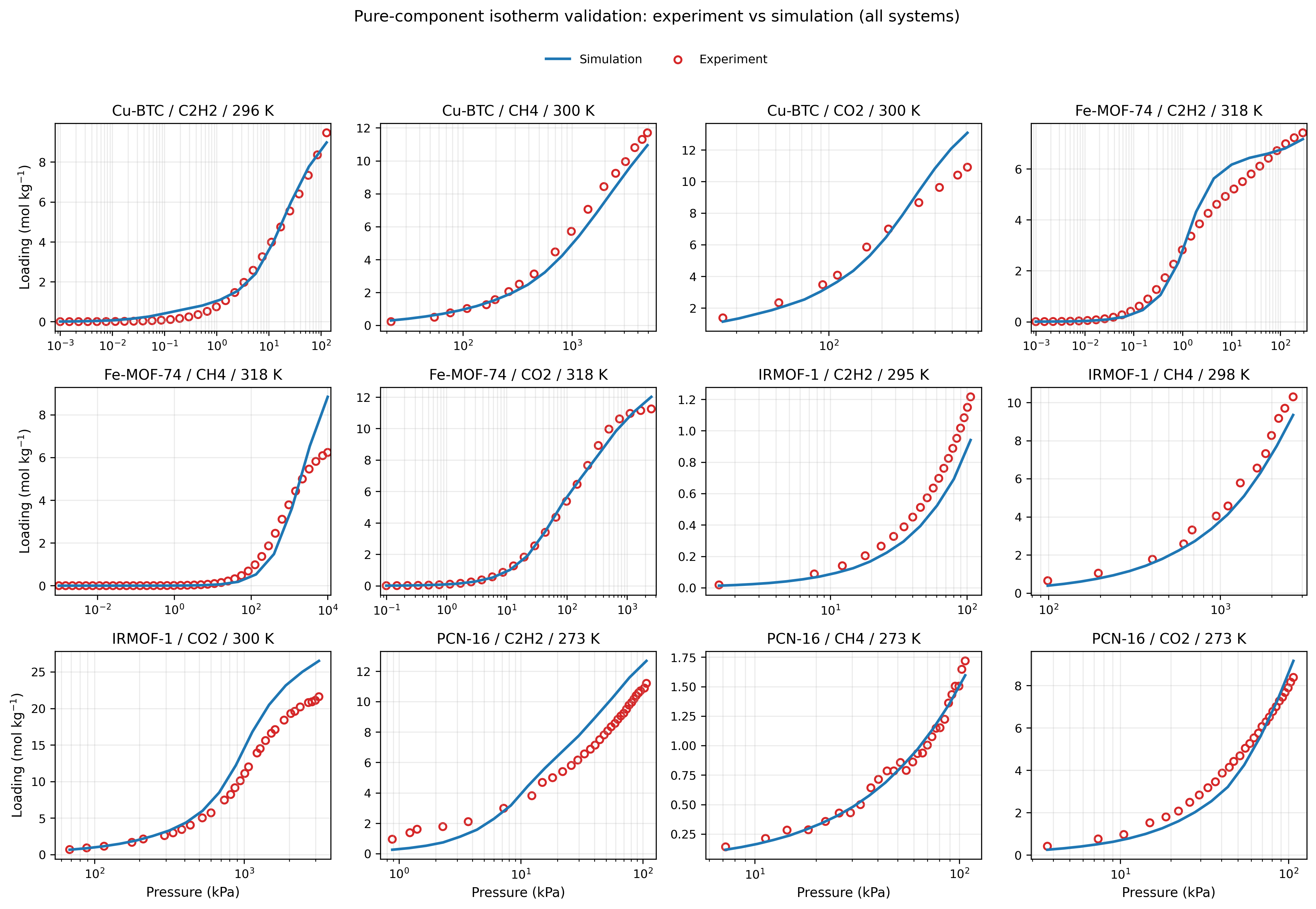}
\caption{Pure-component validation overview from Study~2, multi-agent Run~2, for the four MOF materials and supplied experimental reference data~\cite{SI-luna2019acetylene}. Material-dependent disagreement is not itself a scoring failure.}
\label{SI-fig:si_acetylene_validation}
\end{figure}
\FloatBarrier

\subsubsection{Metal-center effects in M-MOF-74}
Table~\ref{SI-tab:mof74_workflow} defines the 40-point M-MOF-74 workflow across Co, Cu, Fe, Mn, Ni, and Zn. The authoritative experimental conventions use bar for CO$_2$ isotherm pressure and mol~kg$^{-1}$ for loading at 298~K. H$_2$ isotherms use kPa for pressure and mol per metal site for loading at 77~K. For adsorption heats, CO$_2$ loading is expressed as $(N_{\mathrm{CO_2}}/N_M)\times100\%$ and H$_2$ loading as mol per metal site. Heats are in kJ~mol$^{-1}$. CO$_2$ binding uses the closest O--metal distance, whereas H$_2$ binding uses the H$_2$ centre-of-mass--metal distance. Both geometries are evaluated at 298~K.

\begin{table}[htbp]
\centering
\caption{Workflow criteria for metal-center effects in M-MOF-74~\cite{SI-wasik2024impact}.}
\label{SI-tab:mof74_workflow}
\small
\begin{tabularx}{\linewidth}{l X r}
\toprule
\textbf{ID} & \textbf{Expected workflow step} & \textbf{Points} \\
\midrule
M74-W1 & Audit all six materials and the experimental units and definitions. & 3 \\
M74-W2 & Validate CO$_2$ at 298~K and H$_2$ at 77~K for all six metals. & 6 \\
M74-W3 & Validate CO$_2$ and H$_2$ adsorption heats using the correct loading axes. & 4 \\
M74-W4 & Validate CO$_2$ and H$_2$ binding geometries at 298~K. & 4 \\
M74-W5 & Generate and audit 298~K pure CO$_2$ and H$_2$ data and fits for process calculations. & 4 \\
M74-W6 & Compare IAST with representative direct mixture GCMC and judge reliability. & 5 \\
M74-W7 & Evaluate all six feeds at 298~K, 2.5~MPa, void fraction 0.4, velocity 0.006791~m~s$^{-1}$, and length 0.065~m. & 7 \\
M74-W8 & Analyze an equimolar 4~MPa mixture for both species in all six materials. & 4 \\
M74-W9 & Synthesize rankings, limitations, and evidence paths. & 3 \\
\bottomrule
\end{tabularx}
\end{table}
\FloatBarrier

H$_2$ data at 298~K must be generated for the 298~K IAST and process calculations rather than reusing the 77~K validation isotherm. Interpretation should allow tiers or ties, recognize instability caused by very small H$_2$ loadings, and downgrade IAST-derived breakthrough rankings when direct mixture GCMC disagrees. No exact metal order is prescribed.

\begin{figure}[htbp]
\centering
\includegraphics[width=0.98\linewidth,trim=0 0 0 32bp,clip]{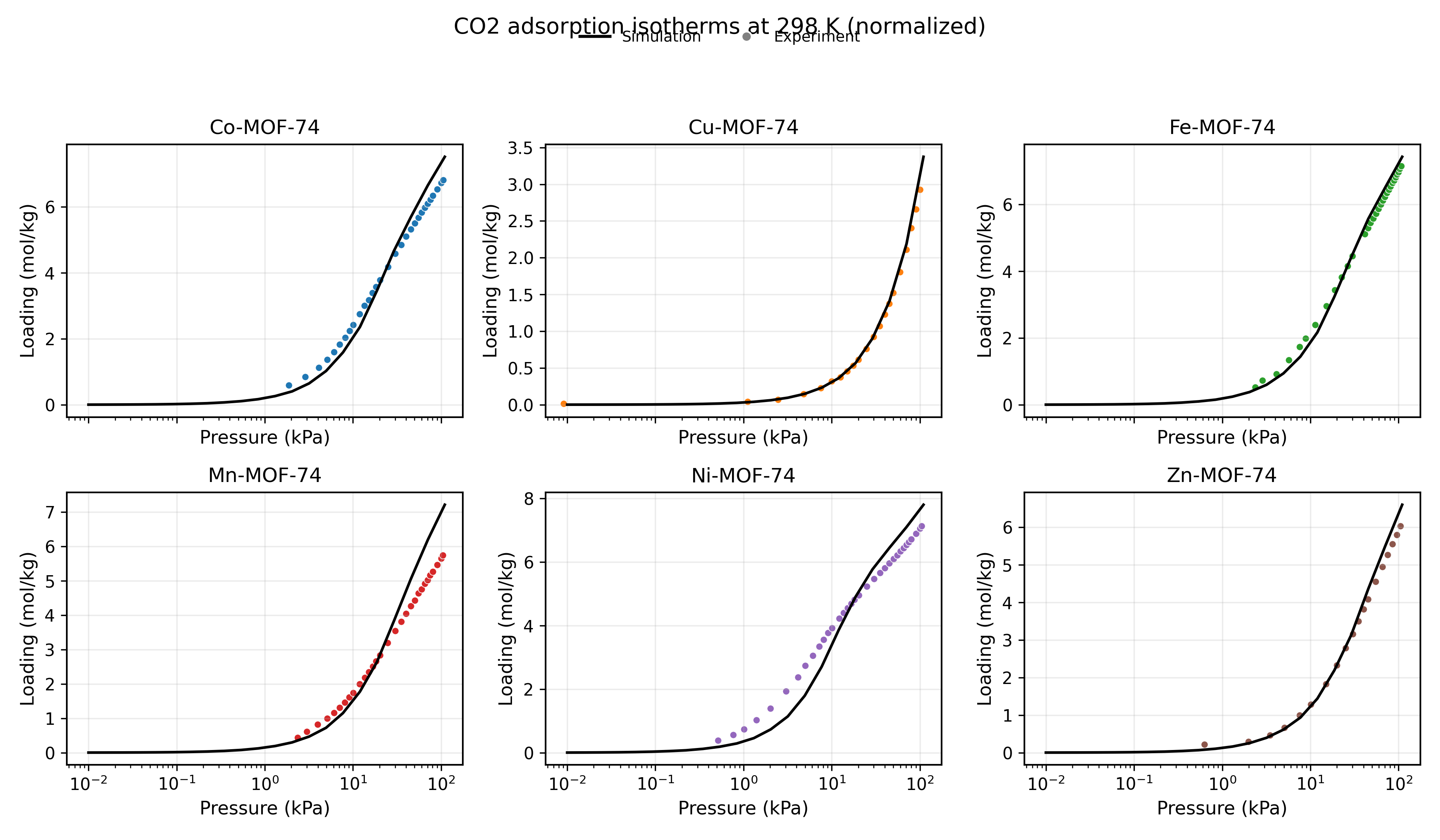}
\caption{CO$_2$ isotherms at 298~K from Study~3, multi-agent Run~2, for the six M-MOF-74 materials. Black lines denote simulations and colored circles the experimental reference data~\cite{SI-wasik2024impact}. Loadings are in mol~kg$^{-1}$. The comparison illustrates both successful reproduction of broad adsorption trends and material-dependent deviations. Numerical agreement is not itself a scoring criterion.}
\label{SI-fig:si_mof74_co2_validation}
\end{figure}
\FloatBarrier

\subsubsection{Olefin/paraffin separation in cation-exchanged zeolites}
The zeolite workflow comprises calculation of Rosenbluth weights, validation of pure propane and propene isotherms and hexane and hexene isobars, mixture adsorption and selectivity, and adsorption-site analysis across cation identities, topologies, and chain lengths~\cite{SI-luna2020pi}. Table~\ref{SI-tab:zeolite_workflow} gives the atomic workflow decomposition. For flexible hydrocarbons, full simulation-validity credit requires the final averaged Widom/Rosenbluth factor at the matching temperature to be propagated into the corresponding GCMC input. Completion of the auxiliary calculation alone is not sufficient when an incorrect value is used downstream.

\begin{table}[htbp]
\centering
\caption{Atomic workflow criteria for the olefin/paraffin zeolite benchmark.}
\label{SI-tab:zeolite_workflow}
\small
\begin{tabularx}{\linewidth}{l X r}
\toprule
\textbf{ID} & \textbf{Expected workflow step} & \textbf{Points} \\
\midrule
ZEO-W1 & Audit structures, cation and restart data, force fields, experimental states, units, and pressure coverage. & 3 \\
ZEO-W2 & Calculate the C$_3$--C$_6$ Rosenbluth prerequisites at 298~K and the C$_6$ values across the isobar temperatures. & 4 \\
ZEO-W3 & Simulate and compare propane and propene isotherms for LTA5A and NaY at 298~K. & 5 \\
ZEO-W4 & Simulate and compare hexane and 1-hexene isobars for LTA5A and NaY at the experimental pressures. & 5 \\
ZEO-W5 & Run equimolar direct-mixture GCMC for all eight frameworks and four chain lengths at 298~K over the pressure range, including 100~kPa. & 9 \\
ZEO-W6 & Calculate and audit selectivity versus pressure and at 100~kPa, including denominator-sensitive points. & 5 \\
ZEO-W7 & Generate and compare 298~K, 100~kPa propene occupancy maps for the five FAU materials. & 5 \\
ZEO-W8 & Synthesize material, pressure, chain-length, cation, validation, and limitation findings in an auditable report. & 4 \\
\midrule
 & \textbf{Total} & \textbf{40} \\
\bottomrule
\end{tabularx}
\end{table}
\FloatBarrier

The supplied cation-position restarts and component-order conventions should be retained where applicable, and supercell replication must be appropriate for the selected cutoff. Exact material or chain-length rankings are not prescribed. Selectivities based on very small paraffin loadings should be treated as denominator-sensitive, and occupancy-map comparisons should distinguish within-framework localization from absolute density. Claims assigning adsorption changes to cation identity or position require direct spatial, energetic, or otherwise controlled evidence. Numerical disagreement with the experimental isotherms or isobars is not itself penalized.

\subsection{Force-Field Extraction Evaluation}
\label{SI-sec:ff_extraction_evaluation}

Force-field extraction is evaluated for parameter correctness and RASPA readiness. The combined score is
\begin{equation}
S_{\mathrm{FF}} = 0.60S_{\mathrm{parameters}} + 0.40S_{\mathrm{readiness}}.
\end{equation}
The score is calculated separately for each repeated extraction and then averaged across repetitions for each case. The benchmark-level score is the arithmetic mean of the eight case scores.

\paragraph{Parameter correctness.}
The extracted parameter set is compared with a reference set derived from the original publications. A parameter is any independent numerical quantity that influences the simulation: Lennard--Jones $\sigma$ and $\epsilon$ are counted separately, as are atomic charges where applicable, molecular bond lengths, and other intramolecular or bonded terms. A parameter is considered correct only when its identifier, interaction form, value, and units exactly match the reference definition. No numerical tolerance is applied. Performance is quantified using an intersection-over-union score,
\begin{equation}
S_{\mathrm{parameters}} = \frac{\left|P_{\mathrm{correct}}\right|}{\left|P_{\mathrm{reference}} \cup P_{\mathrm{predicted}}\right|},
\end{equation}
where $P_{\mathrm{correct}}$ is the set of correctly extracted parameters. Missing, hallucinated, and incorrectly valued parameters reduce the score.

\paragraph{RASPA readiness.}
This metric measures whether the generated force-field files can be used directly in RASPA without manual correction:
\begin{equation}
S_{\mathrm{readiness}} = \frac{N_{\mathrm{executable}}}{N_{\mathrm{generated}}}.
\end{equation}
A generated file is considered non-executable if it produces a RASPA error, violates formatting or bookkeeping requirements, or requires manual editing before a simulation can run.

Table~\ref{SI-tab:ff_scope} lists the eight force-field extraction cases included in the benchmark.

\begin{table}[htbp]
\centering
\caption{Scope of the force-field extraction benchmark.}
\label{SI-tab:ff_scope}
\begin{tabularx}{\linewidth}{l X}
\toprule
\textbf{Case} & \textbf{Source} \\
\midrule
EPM2 CO$_2$ & Harris and Yung~\cite{SI-harris1995carbon} \\
CO$_2$ in zeolites & Garcia-Sanchez et al.~\cite{SI-garcia2009transferable} \\
CO$_2$ in zeolites & Fang et al.~\cite{SI-fang2013first} \\
CO$_2$, N$_2$, O$_2$, and Ar in zeolites & Vujic et al.~\cite{SI-vujic2016transferable} \\
TraPPE-zeo & Bai et al.~\cite{SI-bai2013trappe} \\
Hydrocarbons and permanent gases in zeolites & Findley and Sholl~\cite{SI-findley2021transferable} \\
TAMOF-1 & Gooijer et al.~\cite{SI-gooijer2025tamof} \\
UFF and DREIDING for MOFs & Rappe et al. and Mayo et al.~\cite{SI-rappe1992uff,SI-mayo1990dreiding} \\
\bottomrule
\end{tabularx}
\end{table}

\subsection{Force-Field Fitting Harness}
\label{SI-sec:fitting_harness}

Force-field optimization was controlled through two calls:
\begin{quote}
\small\ttfamily
enter\_fitting\_loop(strategy, objective,\\
\hspace*{2em}stop\_criteria, max\_attempts)\\
complete\_fitting\_phase(decision)
\end{quote}
The controller enforced the sequence parameter selection $\rightarrow$ simulation $\rightarrow$ analysis $\rightarrow$ continuation or termination, recorded the attempt number and declared scientific criteria, and preserved each iteration separately. The allowed attempt limit was 1--50. The Fe-MOF-74/C$_2$H$_2$ campaign used 15 attempts with three candidates per attempt, giving 45 evaluated variants.

Candidate generation followed an adaptive, hypothesis-driven local search. The initial stage varied the explicit Fe--C$_{\mathrm{C2H2}}$ Lennard--Jones cross interaction over $\epsilon=120$--1200~K and $\sigma=2.30$--3.20~\AA. Subsequent stages introduced explicit O1/O2/O3--C$_{\mathrm{C2H2}}$ terms, first as a grouped interaction and then by oxygen type, over the approximate ranges $\epsilon=10$--50~K and $\sigma=2.80$--3.60~\AA. Three candidates were selected near the most promising feasible region after each attempt, using the observed sensitivity of adsorption to $\epsilon$ and binding geometry to $\sigma$.

Candidates were compared by multi-criterion, guardrail-constrained selection rather than a single scalar loss. The criteria comprised Fe--C binding-distance deviations from experimental values of 2.48 and 2.52~\AA\ at 298~K, logarithmic loading errors at 0.055 and 1.696~kPa at 318~K, and the 318~K Widom Henry coefficient. Widom adsorption energy was monitored as a diagnostic and did not determine candidate acceptance or rejection. Fe-MOF-74/CO$_2$ binding geometry and loading at 0.1~kPa served as guardrails, and candidates that lost the open-metal-site CO$_2$ pose were rejected. The loop terminated when both Fe--C distances were within 0.10~\AA\ of experiment, both C$_2$H$_2$ loadings were within a factor of two of experiment, and the CO$_2$ guardrails were retained, or when the 15-attempt limit was reached.

\subsection{Independent Energetic Validation}
\label{SI-sec:energetic_validation}

\begin{figure}[htbp]
\centering
\includegraphics[width=0.72\linewidth]{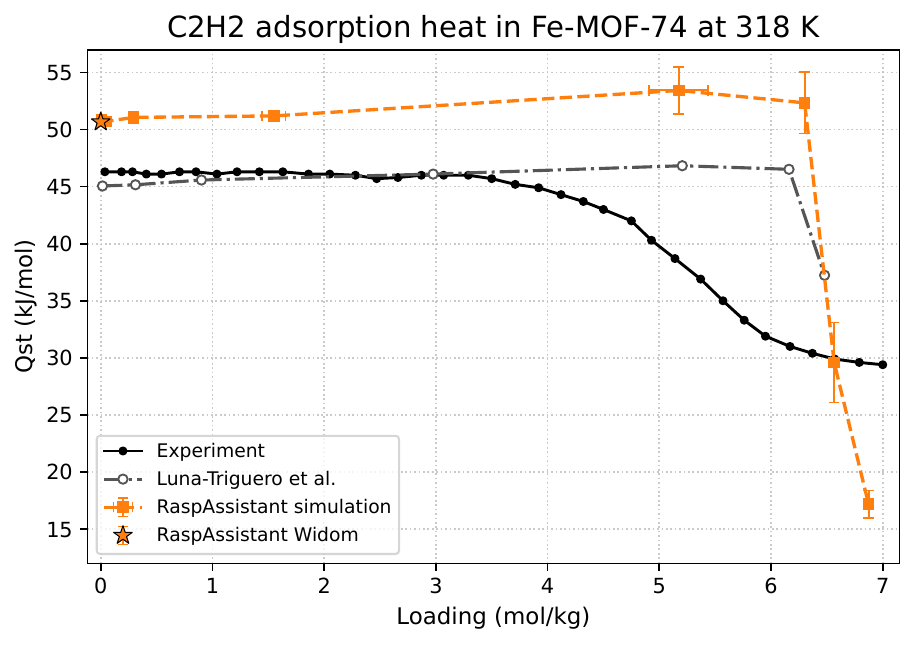}
\caption{Post-fitting energetic assessment for C$_2$H$_2$ in Fe-MOF-74 at 318~K. The experimental curve is from Bloch et al.~\cite{SI-bloch2012hydrocarbon}. The literature-force-field points were digitized from Ref.~\cite{SI-luna2019acetylene}. Experimental heats were unavailable to the fitting agent. RaspAssistant results use the final fitted force field. The zero-loading Widom prediction was monitored without an experimental heat reference or use in selection. Error bars show the reported simulation uncertainties and are smaller than the symbols where not visible.}
\label{SI-fig:si_acetylene_qst}
\end{figure}

Experimental adsorption-heat data were unavailable to the fitting agent and were not used as optimization targets. The fit used the C$_2$H$_2$ isotherm and binding geometry. Predicted Widom adsorption energy was monitored without guiding candidate selection or comparison with experimental heats. The literature model was also validated against adsorption-heat data~\cite{SI-luna2019acetylene}. Figure~\ref{SI-fig:si_acetylene_qst} compares the final fitted model with experimental heats after fitting, providing an independent energetic assessment. The fitted model reproduces the broad low-loading energy scale and the decrease toward saturation. The decrease occurs more sharply than in experiment, while the literature force field follows the low-loading measurements more closely but likewise departs from the experimental curve over part of the high-loading range. Including adsorption-heat observables in future fitting campaigns could further constrain interaction models that otherwise reproduce similar adsorption isotherms and binding geometries.

\FloatBarrier

\clearpage

\providecommand{\latin}[1]{#1}
\makeatletter
\providecommand{\doi}
  {\begingroup\let\do\@makeother\dospecials
  \catcode`\{=1 \catcode`\}=2 \doi@aux}
\providecommand{\doi@aux}[1]{\endgroup\texttt{#1}}
\makeatother
\providecommand*\mcitethebibliography{\thebibliography}
\csname @ifundefined\endcsname{endmcitethebibliography}
  {\let\endmcitethebibliography\endthebibliography}{}

\end{document}